# Loss function based second-order Jensen inequality and its application to particle variational inference

Futoshi Futami*[1], Tomoharu Iwata[1], Naonori Ueda[1,3], Issei Sato[2], and Masashi Sugiyama[3,2]

[1]Communication Science Laboratories, NTT, Kyoto, Japan
[2]The University of Tokyo, Tokyo, Japan
[3]RIKEN, Tokyo, Japan

June 11, 2021

## Abstract

Bayesian model averaging, obtained as the expectation of a likelihood function by a posterior distribution, has been widely used for prediction, evaluation of uncertainty, and model selection. Various approaches have been developed to efficiently capture the information in the posterior distribution; one such approach is the optimization of a set of models simultaneously with interaction to ensure the diversity of the individual models in the same way as ensemble learning. A representative approach is particle variational inference (PVI), which uses an ensemble of models as an empirical approximation for the posterior distribution. PVI iteratively updates each model with a repulsion force to ensure the diversity of the optimized models. However, despite its promising performance, a theoretical understanding of this repulsion and its association with the generalization ability remains unclear. In this paper, we tackle this problem in light of PAC-Bayesian analysis. First, we provide a new second-order Jensen inequality, which has the repulsion term based on the loss function. Thanks to the repulsion term, it is tighter than the standard Jensen inequality. Then, we derive a novel generalization error bound and show that it can be reduced by enhancing the diversity of models. Finally, we derive a new PVI that optimizes the generalization error bound directly. Numerical experiments demonstrate that the performance of the proposed PVI compares favorably with existing methods in the experiment.

## 1 Introduction

Bayesian model averaging (BMA) has been widely employed for prediction, evaluation of uncertainty, and model selection in Bayesian inference. BMA is obtained as the expectation of a likelihood function by a posterior distribution and thus it contains information of each model drawn from the posterior distribution [18]. Since estimating the posterior distribution is computationally difficult in practice, various approximations have been developed to efficiently capture the diversity in the posterior distribution [18, 1, 2].

*futoshi.futami.uk@hco.ntt.co.jp

One of these recently proposed approaches involves optimizing a set of models simultaneously with interaction to ensure the diversity of the individual models, similar to ensemble learning. One notable example is particle variational inference (PVI) [14, 24], which uses an ensemble as an empirical approximation for the posterior distribution. Such PVI methods have been widely employed in variational inference owing to their high computational efficiency and flexibility. They iteratively update the individual models and the update equations contain the gradient of the likelihood function and the repulsion force that disperses the individual models. Thanks to this repulsion term, the obtained ensemble can appropriately approximate the posterior distribution. When only one model is used in PVI, the update equation is equivalent to that of the maximum a posteriori (MAP) estimation. Other methods have been developed apart from PVI, especially for latent variable models, which have introduced regularization to the MAP objective function to enforce the diversity in the ensemble. A notable example of such methods is the determinantal point process (DPP) [25].

Despite successful performances of these methods in practice [19, 15, 14, 24, 5, 23], a theoretical understanding of the repulsion forces remains unclear. Some previous studies considered PVI as a gradient flow in Wasserstein space with an infinite ensemble size [13, 11] and derived the convergence theory. However, an infinite ensemble size is not a practical assumption and no research has been conducted to analyze the repulsion force related to the generalization.

BMA can be regarded as a special type of ensemble learning [22], and recent work has analyzed the diversity of models in ensemble learning in light of the PAC-Bayesian theory [16]. They reported that the generalization error is reduced by increasing the variance of *the predictive distribution*. However, the existing posterior approximation methods, such as PVI and DPPs, enhance the diversity with the repulsion of the parameters or models rather than the repulsion of the predictive distribution. We also found that the analysis in a previous work [16] cannot be directly extended to the repulsion of parameters or loss functions (see Appendix D). In addition, when the variance of the predictive distribution is included in the objective function in the variational inference, the obtained model shows large epistemic uncertainty, which hampers the fitting of each model to the data (see Section 5).

Based on these findings, this study aims to develop a theory that explains the repulsion forces in PVI and DPPs and elucidates the association of the repulsion forces with the generalization error. To address this, we derive the novel second-order Jensen inequality and connect it to the PAC-Bayesian generalization error analysis. Our second-order Jensen inequality includes the information of the variance of loss functions. Thanks to the variance term, our bound is tighter than the standard Jensen inequality. Then, we derive a generalization error bound that includes the repulsion term, which means that enhancing the diversity is necessary to reduce the generalization error. We also show that PVI and DPPs can be derived from our second-order Jensen inequality, and indicate that these methods work well from the perspective of the generalization error. However, since these existing methods do not minimize the generalization error upper bound, there is still room for improvement. In this paper, we propose a new PVI that directly minimize the generalization error upper bound and empirically demonstrate its effectiveness.

Our contributions are summarized as follows:

1. We derive a novel second-order Jensen inequality that inherently includes the variance of loss functions. Thanks to this variance term, our second-order Jensen inequality is tighter than the standard Jensen inequality. We then show that enhancing the diversity is important for reducing the generalization error bound in light of PAC-Bayesian analysis.

2. From our second-order Jensen inequality, we derive the existing PVI and DPPs. We demonstrate that these methods work well even at a finite ensemble size, since their objective functions includes valid diversity enhancing terms to reduce the generalization error.

3. We propose a new PVI that minimizes the generalization error bound directly. We numerically demonstrate that the performance of our PVI compares favorably with existing methods.

## 2 Background

In this section, we briefly review PVI, DPPs, and PAC-Bayesian analysis.

### 2.1 Particle variational inference

Assume that training dataset $\mathcal{D} = (x_1, \ldots, x_D)$ is drawn independently from unknown data generating distribution $\nu(x)$. Our goal is to model $\nu(x)$ by using a parametrized model $p(x|\theta)$, where $\theta \in \Theta \subset \mathbb{R}^d$. We express $p(\mathcal{D}|\theta) = \sum_{d=1}^{D} \ln p(x_d|\theta)$ and assume a probability distribution over parameters. In Bayesian inference, we incorporate our prior knowledge or assumptions into a prior distribution $\pi(\theta)$. This is updated to a posterior distribution $p(\theta|\mathcal{D}) \propto p(\mathcal{D}|\theta)\pi(\theta)$, which incorporates the observation $\mathcal{D}$. Let us consider the approximation of $p(\theta|\mathcal{D})$ with $q(\theta)$. We predict a new data point by a predictive distribution $p(x) = \mathbb{E}_{q(\theta)} p(x|\theta)$, where $\mathbb{E}$ denotes the expectation. This expectation over the posterior is often called BMA [18].

Assume that we draw $N$ models from the posterior distribution and calculate BMA. We denote those drawn models as an empirical distribution $\rho_{\mathrm{E}}(\theta) = \frac{1}{N} \sum_{i=1}^{N} \delta_{\theta_i}(\theta)$, where $\delta_{\theta_i}(\theta)$ is the Dirac distribution that has a mass at $\theta_i$. We also refer to these $N$ models as particles. The simplest approach to obtain these $N$ particles is MAP estimate that updates parameters independently with gradient descent (GD) as follows [24]:

$$\theta_i^{\mathrm{new}} \leftarrow \theta_i^{\mathrm{old}} + \eta \partial_\theta \ln p(D|\theta_i^{\mathrm{old}}) \pi(\theta_i^{\mathrm{old}}), \tag{1}$$

where $\eta \in \mathbb{R}^+$ is a step size. In BMA, we are often interested in the multimodal information of the posterior distribution. In such a case, MAP estimate is not sufficient because $N$ optimized particles do not necessarily capture the appropriate diversity of the posterior distribution. Instead, particle variational inference (PVI) [14, 24] approximates the posterior through iteratively updating the empirical distribution by interacting them with each other:

$$\theta_i^{\mathrm{new}} \leftarrow \theta_i^{\mathrm{old}} + \eta v(\{\theta_{i'}^{\mathrm{old}}\}_{i'=1}^N), \tag{2}$$

Table 1: Particle variational inference methods. $I$ is an $N \times N$ identity matrix and $K_{i,j} := K(\theta_i, \theta_j)$.

| Methods | $v(\theta)$ |
|---------|-------------|
| SVGD[14] | $\frac{1}{N}\sum_{j=1}^{N} K_{ij}\partial_{\theta_j} \log p(\mathcal{D}|\theta_j)\pi(\theta_j) + \partial_{\theta_j} K_{ij}$ |
| w-SGLD[3] | $\partial_{\theta_i} \log p(\mathcal{D}|\theta_i)\pi(\theta_i) + \sum_{j=1}^{N} \frac{\partial_{\theta_j} K_{ij}}{\sum_{k=1}^{N} K_{jk}} + \frac{\sum_{j=1}^{N} \partial_{\theta_j} K_{ji}}{\sum_{k=1}^{N} K_{ik}}$ |
| GFSD[3] | Sum of SVGD and w-SGLD |
| GFSF[24] | $\partial_{\theta_i} \log p(\mathcal{D}|\theta_i)\pi(\theta_i) + \frac{1}{N}\sum_{j=1}^{N}((K+cI)^{-1})_{ij}\partial_{\theta_j} K_{ij}$ |

where $v(\{\theta\})$ is the update direction and explicit expressions are summarized in Table 1. Basically, $v$ is composed of the gradient term and the repulsion term. In Table 1, the repulsion terms contain the derivative of the kernel function $K$, and the Gaussian kernel [20] is commonly used. When the bandwidth of $K$ is $h$, the repulsion term is expressed as $\partial_{\theta_i} K(\theta_i, \theta_j) = -h^{-2}(\theta_i - \theta_j)e^{-(2h^2)^{-1}\|\theta_i - \theta_j\|^2}$, where $\|\cdot\|$ denotes the Euclidean norm. We refer to this as a parameter repulsion. Note that the repulsion term depends on the distance between particles, and the closer they are, the stronger force is applied. This moves $\theta_i$ away from $\theta_j$, and thus particles tend not to collapse to a single mode.

For over-parametrized models such as neural networks, since the repulsion in the parameter space is not enough for enhancing the diversity, function space repulsion force for supervised tasks was developed [24]. We call it function space PVI (f-PVI). Pairs of input-output data are expressed as $\mathcal{D} = \{(x, y)\}$. We consider the model $p(y|x, \theta) = p(y|f(x; \theta))$ where $f(x; \theta)$ is a $c$-dimensional output function parametrized by $\theta$ and $x$ is an input. Furthermore, we consider the distribution over $f$ and approximate it by a size-$N$ ensemble of $f$, which means that we prepare $N$ parameters (particles) $\{\theta_i\}_{i=1}^{N}$. We define $f_i(x) := f(x; \theta_i)$. When we input the minibatch with size $b$ into the model, we express it as $f_i(\boldsymbol{x}_{1:b}) = (f_i(x_1), \ldots, f_i(x_b)) \in \mathbb{R}^{cb}$. Then the update equation is given as

$$\theta_i^{\text{new}} \leftarrow \theta_i^{\text{old}} + \eta \frac{\partial f_i(\boldsymbol{x}_{1:b})}{\partial \theta_i}\Big|_{\theta_i = \theta_i^{\text{old}}} v(\{f_i(\boldsymbol{x}_{1:b})\}_{i=1}^{N}), \tag{3}$$

where $v(\{f_i\})$ is obtained by replacing $\ln p(\mathcal{D}|\theta)\pi(\theta)$ with $(D/b)\sum_{d=1}^{b} \ln p(x_d|\theta)\pi(f)$, where $\pi(f)$ is a prior distribution over $f$ and the Gram matrix $K(\theta_i, \theta_j)$ is replaced with $K(f_i^b, f_j^b)$ in Table 1. See appendix A.1 for details. Then, f-PVI modifies the loss signal so that models are diverse. We refer to the repulsion term of f-PVI as a model repulsion. We express $f_i^b := f_i(\boldsymbol{x}_{1:b})$ for simplicity. When we use the Gaussian kernel, the model repulsion is expressed as

$$\partial_{\theta_i} K(f_i^b, f_j^b) = -h^{-2}(f_i^b - f_j^b)e^{-\|f_i^b - f_j^b\|^2/(2h^2)}\partial_{\theta_i} f_i^b. \tag{4}$$

Thus, the model repulsion pushes model $f_i$ away from $f_j$.

## 2.2 Regularization based methods and determinantal point processes

Another common approach for enhancing the diversity for latent variable models is based on regularization. A famous example is the determinantal point process (DPP) [25], in which we maximize

$$\mathbb{E}_{\rho_E} \ln p(\mathcal{D}|\theta)\pi(\theta) + \ln \det K, \tag{5}$$

where $K$ is the kernel Gram matrix defined by $K_{i,j} = K(\theta_i, \theta_j)$. This log-determinant term is essentially a repulsion term that enhances the diversity in the parameter space.

### 2.3 PAC-Bayesian theory

Here, we introduce PAC-Bayesian theory [7]. We define the generalization error as the cross-entropy:

$$\text{CE} := \mathbb{E}_{\nu(x)}[-\ln \mathbb{E}_{q(\theta)} p(x|\theta)], \tag{6}$$

which corresponds to the Kullback-Leibler (KL) divergence. Our goal is to find $q(\theta)$ that minimizes the above CE. In many Bayesian settings, we often minimize not CE but a surrogate loss [16] that is obtained by the Jensen inequality:

$$\mathbb{E}_{\nu(x)}[-\ln \mathbb{E}_{q(\theta)} p(x|\theta)] \leq \mathbb{E}_{\nu(x),q(\theta)}[-\ln p(x|\theta)]. \tag{7}$$

Since the data generating distribution is unknown, we approximate it with a training dataset as $\mathbb{E}_{\nu(x),q(\theta)}[-\ln p(x|\theta)] \approx \mathbb{E}_{q(\theta)} \frac{1}{D} \sum_{d=1}^{D}[-\ln p(x_d|\theta)]$. The PAC-Bayesian generalization error analysis provides the probabilistic relation for this approximation as follows:

**Theorem 1.** *[7] For any prior distribution $\pi$ over $\Theta$ independent of $\mathcal{D}$ and for any $\xi \in (0,1)$ and $c > 0$, with probability at least $1-\xi$ over the choice of training data $\mathcal{D} \sim \nu^{\otimes D}(x)$, for all probability distributions $q$ over $\Theta$, we have*

$$\mathbb{E}_{\nu(x),q(\theta)}[-\ln p(x|\theta)] \leq \mathbb{E}_{q(\theta)} \frac{1}{D} \sum_{d=1}^{D}[-\ln p(x_d|\theta)] + \frac{\text{KL}(q,\pi) + \ln \xi^{-1} + \Psi_{\pi,\nu}(c,D)}{cD}, \tag{8}$$

where $\Psi_{\pi,\nu}(c,D) := \ln \mathbb{E}_\pi \mathbb{E}_{\mathcal{D} \sim \nu^{\otimes D}(x)} \exp[cD(-\mathbb{E}_{\nu(x)} \ln p(x|\theta) + D^{-1} \sum_{d=1}^{D} \ln p(x_d|\theta))]$.

The Bayesian posterior is the minimizer of the right-hand side of the PAC-Bayesian bound when $c = 1$. Recently, the PAC-Bayesian bound has been extended so that it includes the diversity term [16]. Under the same assumptions as Theorem 1 and for all $x, \theta, p(x|\theta) < \infty$, we have

$$\text{CE} \leq -\mathbb{E}_{\nu,q}[\ln p(x|\theta) + V(x)]$$
$$\leq -\mathbb{E}_q \frac{1}{D} \sum_{d=1}^{D}[\ln p(x_d|\theta) + V(x_d)] + \frac{\text{KL}(q,\pi) + \frac{\ln \xi^{-1} + \Psi'_{\pi,\nu}(c,D)}{2}}{cD}, \tag{9}$$

where $V(x) := (2 \max_\theta p(x|\theta)^2)^{-1} \mathbb{E}_{q(\theta)}\left[(p(x|\theta) - \mathbb{E}_{q(\theta)} p(x|\theta))^2\right]$ is the variance of the predictive distribution and $\Psi'_{\pi,\nu}(c,D)$ is the modified constant of $\Psi_{\pi,\nu}(c,D)$ (see Appendix A.2 for details). A similar bound for ensemble learning, that is, $q(\theta)$ as an empirical distribution, was also previously proposed [16] (see Appendix A.3). This bound was derived directly from the second-order Jensen inequality derived in another work [12]. Furthermore, the diversity comes from the variance of the predictive distribution, which is different from PVI and DPPs because their repulsion is in the parameter or model space. Note that we cannot directly change the variance of the predictive distribution to that of PVI or DPPs because it requires an inequality that is contrary to the Jensen inequality. We also found that directly optimizing the upper bound of Eq.(9), referred to as $\text{PAC}_\text{E}^2$, results in a too large variance of the predictive distribution which is too pessimistic for supervised learning tasks (see Section 5).

# 3 Method

Here, we derive our novel second-order Jensen inequality based on the variance of loss functions and then derive a generalization error bound. Then, we connect our theory with existing PVI and DPPs.

## 3.1 A novel second-order Jensen inequality

First, we show the second-order equality, from which we derive our second-order Jensen inequality.

**Theorem 2.** *Let $\psi$ be a twice differentiable monotonically increasing concave function on $\mathbb{R}^+$, $Z$ be a random variable on $\mathbb{R}^+$ that satisfies $\mathbb{E}Z^2 < \infty$, and its probability density be $p_Z(z)$. Define a constant $\mu := \psi^{-1}(\mathbb{E}[\psi(Z)])$. Then, we have*

$$\mathbb{E}[Z] = \mu - \left(2\frac{d\psi(\mu)}{dz}\right)^{-1} \int_{\mathbb{R}^+} \left[\frac{d^2\psi(c(z))}{dz^2}(z-\mu)^2\right] p_Z(z)dz, \qquad (10)$$

*where $c(z)$ is a constant between $z$ and $\mu$ that is defined from the Taylor expansion (see Appendix B.1 for details).*

*Proof sketch:* There exists a constant $c(z)$ between $z$ and $\mu$ s.t. $\psi(z) = \psi(\mu) + \frac{d\psi(\mu)}{dz}(z-\mu) + \frac{1}{2}\frac{d^2\psi(c(z))}{dz^2}(z-\mu)^2$ from the Taylor expansion. Then we take the expectation. Full proof is given in Appendix B.1. □

This theorem states the deviation of $\mathbb{E}Z$ from $\mu$ when $\psi$ is applied to $Z$. By setting $\psi(\cdot) = \ln(\cdot)$ and $Z = p(x|\theta)$ and applying ln to both hand sides of Eq.(10), we have the following equality:

**Corollary 1.** *If for all $x$ and $\theta$, $p(x|\theta) < \infty$, we have*

$$\mathbb{E}_{q(\theta)} \ln p(x|\theta)$$
$$= \ln \mathbb{E}_{q(\theta)} p(x|\theta) - \ln\left(1 + \mathbb{E}_{q(\theta)}(2g(\theta,x)^2)^{-1}(e^{\ln p(x|\theta)} - e^{\mathbb{E}_{q(\theta)} \ln p(x|\theta)})^2\right), \quad (11)$$

*where $g(\theta, x)$ is a constant between $p(x|\theta)$ and $e^{\mathbb{E}_{q(\theta)} \ln p(x|\theta)}$ that is defined from the Taylor expansion (see Appendix B.2 for details).*

**Remark 1.** *Recall that the standard Jensen inequality is $\mathbb{E}_q \ln p(x|\theta) \leq \ln \mathbb{E}_q p(x|\theta)$, and its gap is called the Jensen gap. In Eq.(11), the second term of the right-hand side is always positive. Thus, this term corresponds to the Jensen gap. Remarkably when we use the standard Jensen inequality, this information is lost. We clarify the meaning of this term below. Also note that our second-order equality is different from those of the previous works [16, 12] (see Appendix D for details).*

Next, we show our first main result, loss function based second-order Jensen inequality:

**Theorem 3.** *Under the same assumption as Corollary 1,*

$$\mathbb{E}_{q(\theta)} \ln p(x|\theta) \leq \ln \mathbb{E}_{q(\theta)} p(x|\theta) - \underbrace{\mathbb{E}_{q(\theta)}\left(\frac{\ln p(x|\theta) - \mathbb{E}_{q(\theta)} \ln p(x|\theta)}{2h(x,\theta)}\right)^2}_{:= \mathrm{R}(x,h)}, \qquad (12)$$

where

$$h(x,\theta)^{-2} = \exp\left(\ln p(x|\theta) + \mathbb{E}_{q(\theta)} \ln p(x|\theta) - 2\max_{\theta} \ln p(x|\theta)\right). \quad (13)$$

*Proof sketch:* Apply $\sqrt{\alpha\beta} \leq \frac{\alpha-\beta}{\ln\alpha-\ln\beta}$ for any $\alpha, \beta > 0$ to Eq.(11). Full proof is given in Appendix B.3. □

**Remark 2.** *$R$ is the weighted variance of loss functions, and it is always positive. Thus, this inequality is always tighter than the Jensen inequality, and the equality holds if the weighted variance is zero. Compared to the results of the previous works [16, 12] that used the predictive variance in the inequality, our bound focuses on the variance of loss functions. We refer to our repulsion term $R$ as a loss repulsion.*

Then, by rearranging Eq.(12) and taking the expectation, we have the following inequality:

$$\text{CE} \leq -\mathbb{E}_{q(\theta),\nu(x)}[\ln p(x|\theta)] - \mathbb{E}_{\nu(x)} R(x,h) \leq -\mathbb{E}_{q(\theta),\nu(x)}[\ln p(x|\theta)]. \quad (14)$$

Using this inequality, we obtain the second-order PAC-Bayesian generalization error bound:

**Theorem 4.** *(See Appendix B.4 for the complete statement) Under the same notation and assumptions as Theorems 1 and 3, with probability at least $1-\xi$, we have*

$$\text{CE} \leq -\mathbb{E}_{\nu,q}[\ln p(x|\theta) + R(x,h)]$$
$$\leq -\mathbb{E}_q \frac{1}{D}\sum_{d=1}^{D}[\ln p(x_d|\theta) + R(x_d, h_m)] + \frac{\text{KL}(q,\pi) + \frac{\ln\xi^{-1}+\Psi''_{\pi,\nu}(c,D)}{3}}{cD}, \quad (15)$$

*where $\Psi''_{\pi,\nu}(c,D)$ is the modified constant of $\Psi_{\pi,\nu}(c,D)$ and $R(x,h_m)$ is $R(x,h)$ in Eq.(12) replacing $h(x,\theta)^{-2}$ of Eq.(13) with*
$h_m(x,\theta)^{-2} = \exp\left(\ln p(x|\theta) + \min_{\theta} \ln p(x|\theta) - 2\max_{\theta} \ln p(x|\theta)\right).$

*Proof sketch.* We express $\mathbb{E}_{q(\theta)}[\ln p(x|\theta) + R(x,h_m)]$ as $\mathbb{E}_{q(\theta)q(\theta')q(\theta'')} L(x,\theta,\theta',\theta'')$ where $L(x,\theta,\theta',\theta'') := \ln p(x|\theta) + (2h_m(x,\theta))^{-2}(\ln p(x|\theta)^2 - 2\ln p(x|\theta) \ln p(x|\theta') + \ln p(x|\theta') \ln p(x|\theta''))$. Then, we apply the same proof technique as Theorem 1 [7] to the loss function $L(x,\theta,\theta',\theta'')$ with $\lambda = 3cD$. Full proof is given in Appendix B.4. □

**Remark 3.** *To reduce the upper bound of the generalization error, Eq.(15), we need to control the trade-off between the data fitting term of the negative log-likelihood and enhancing the diversity of the models based on the loss repulsion term $R$.*

**Remark 4.** *In the definition of $h_m$, $\min_{\theta} \ln p(x|\theta)$ is too pessimistic in some cases. If we additionally assume that there exists a positive constant $M$ s.t. $\mathbb{E}_{q(\theta)}[\ln p(x|\theta)]^2 < M < \infty$, we can replace $\min_{\theta} \ln p(x|\theta)$ with $\text{Median}_{\theta}(\ln p(x|\theta)) - M^{1/2}$ in $h_m$ in Theorem 4 (see Appendix B.4.1).*

Compared to Eq.(9), our bound focuses on the variance of loss functions, which has a direct connection to the repulsion of PVI and DPPs (see Section 3.2). Furthermore, we show that optimizing the upper bound in Eq.(15) shows competitive performance with the existing state-of-the-art PVI (see Section 5). Note that this inequality is not restricted to the case where $q(\theta)$ is a set of parameters. We can also use this for parametric variational inference [18].

## 3.2 Diversity in ensemble learning and connection to existing methods

In the following, we focus on the ensemble setting and use a finite set of parameters as $\rho_{\mathrm{E}}(\theta) := \frac{1}{N} \sum_{i=1}^{N} \delta_{\theta_i}(\theta)$ and discuss the relationship of our theory and existing methods. We show the summary of the relationships in Appendix H.

### 3.2.1 Covariance form of the loss repulsion

To emphasize the repulsion between models, we upper-bound Eq.(12) using the covariance:

**Theorem 5.** *Under the same assumption as Corollary 1, we have*

$$\mathbb{E}_{\rho_{\mathrm{E}}(\theta)} \ln p(x|\theta)$$
$$\leq \ln \mathbb{E}_{\rho_{\mathrm{E}}(\theta)} p(x|\theta) - \frac{1}{2(2h_w(x,\theta))^2 N^2} \sum_{i,j=1}^{N} (\ln p(x|\theta_i) - \ln p(x|\theta_j))^2, \quad (16)$$

*where* $h_w(x,\theta)^{-2} = \exp\left(\min_i \ln p(x|\theta_i) + \frac{1}{N} \sum_{i=1} \ln p(x|\theta_i) - 2\max_j \ln p(x|\theta_j)\right)$.

See Appendix B.5 for the proof. We can also show a generalization error bound like Theorem 4 for the ensemble learning setting (see Appendix C.1). In existing PVI and DPPs, the repulsion is based not on the loss function but the parameters or models, as seen in Section 2. We derive the direct connection between our loss repulsion and the model and parameter repulsion below.

### 3.2.2 Relation to w-SGLD and model repulsion

First, from Eq.(16), we derive the direct connection to w-SGLD, which is a kind of PVI introduced in Section 2. Let us define an $N \times N$ kernel Gram matrix $G$ whose $(i,j)$ element is defined as

$$G_{ij} := \exp\left(-(8h_w^2)^{-1} (\ln p(x|\theta_i) - \ln p(x|\theta_j))^2\right). \quad (17)$$

Applying the Jensen inequality to Eq.(16), we obtain

$$\ln \mathbb{E}_{\rho_{\mathrm{E}}(\theta)} p(x|\theta) \geq \mathbb{E}_{\rho_{\mathrm{E}}(\theta)} \ln p(x|\theta) - \frac{1}{N} \sum_{i=1}^{N} \ln \sum_{j=1}^{N} \frac{G_{ij}}{N} \geq \mathbb{E}_{\rho_E(\theta)} \ln p(x|\theta). \quad (18)$$

This is tighter than the standard Jensen inequality. To derive the relation to w-SGLD, we optimize the middle part of Eq.(180) by gradient descent. We express $L(\theta_i) := \ln p(x|\theta_i)$ and do not consider the dependency of $h_w$ on $\theta$ for simplicity. By taking the partial derivative with respect to $\theta_i$, we have

$$\partial_{\theta_i} \ln p(x|\theta_i) + \left( \sum_{j=1}^{N} \frac{\partial_{L(\theta_j)} G_{ij}}{\sum_{k=1}^{N} G_{jk}} + \frac{\sum_{j=1}^{N} \partial_{L(\theta_j)} G_{ji}}{\sum_{k=1}^{N} G_{ik}} \right) \partial_{\theta_i} L(\theta_i). \quad (19)$$

The second term corresponds to the repulsion term, which is equivalent to that of w-SGLD shown in Table 1. The difference is that our Gram matrix $G$ in Eq.(19) depends on the loss function rather than the parameter or model. Using the mean value theorem, it is easy to verify that there exists a constant $C$ such that $\|\ln p(x|\theta_i) - \ln p(x|\theta_j)\|^2 = \|C(\theta_i - \theta_j)\|^2$ (see Appendix E for details), thus we can easily transform the loss repulsion to the parameter or model repulsion.

However, since we cannot obtain the explicit expression of the constant $C$, it is difficult to understand the intuitive relation between our loss repulsion and the parameter or model repulsion. Instead, here we directly calculate $\partial_{\theta_i} G_{i,j}$ and discuss the relation. Due to the space limitation, we only show the relation to the model repulsion in the regression task of f-PVI. See Appendix E for the complete statement including the classification setting of f-PVI and the parameter repulsion of PVI. Following the setting in Section 2.1, for a regression problem, we assume that $p(y|f(x;\theta))$ is the Gaussian distribution with unit variance for simplicity. We define $L(f_i) := \ln p(y|f(x;\theta_i))$ and $dL_{ij} := \partial_{f_i} L(f_i) + \partial_{f_j} L(f_j)$. The derivative of the Gram matrix $G$ is expressed as

$$\partial_{\theta_i} G_{ij} = -(\underbrace{(f_i - f_j)\|dL_{ij}\|^2}_{i)} + \underbrace{\partial_{f_i} L(f_i) dL_{ij}\|f_i - f_j\|^2}_{ii)})(4h_w)^{-2} G_{ij} \partial_{\theta_i} f_i. \quad (20)$$

The first term $i)$ corresponds to the model repulsion of f-PVI shown in Eq.(29) and the second term $ii)$ is the correction term based on the loss function. Thus, our loss repulsion can be translated to the model repulsion of f-PVI plus the correction term using the loss function.

In conclusion, we have confirmed that w-SGLD is directly related to our theory. For Eq.(180), we can also derive a generalization error bound like Theorem 4 (see Appendix C.2 for details). This means that w-SGLD controls the trade-off between the model fitting and enhancing the diversity by optimizing the generalization error bound. This explains the reason why w-SGLD still works well even with a finite ensemble size.

### 3.2.3 Relation to other PVIs

Here, we derive other PVI and DPPs from our theory. First, we derive GFSF shown in Table 1. We express the identity matrix with size $N$ as $I$. We obtain the following upper bound from Eq.(16):

**Theorem 6.** *Under the same assumption as Corollary 1, there exists a positive constant $\epsilon$, such that*

$$\ln \mathbb{E}_{\rho_E(\theta)} p(x|\theta) \geq \mathbb{E}_{\rho_E(\theta)} \ln p(x|\theta) - \frac{2}{\tilde{h} N} \ln \det(\epsilon I + K) + \frac{2}{\tilde{h} N} \geq \mathbb{E}_{\rho_E(\theta)} \ln p(x|\theta), \quad (21)$$

*where $K$ is an $N \times N$ kernel Gram matrix of which $(i, j)$ element is defined as*

$$K_{ij} := \exp\left(-\tilde{h} \ln N (4h_w)^{-2} \left(\ln p(x|\theta_i) - \ln p(x|\theta_j)\right)^2\right), \quad (22)$$

*and $\tilde{h}$ is a constant that is selected to satisfy the relation $\sum_j K_{ij} < N^2 - \epsilon$ for any $i$.*

The proof is shown in Appendix B.6. This is tighter than the standard Jensen inequality.

**Remark 5.** *If all of the parameters $\{\theta_i\}_{i=1}^N$ are not exactly the same, there exists a constant $\tilde{h}$ that satisfies $\sum_j K_{ij} < N - \epsilon$. If all of the parameters $\{\theta_i\}$ are the same, $\sum_j K_{ij} = N$ holds. In such a case, the repulsion term becomes $0$, and thus we do not need to tune the bandwidth.*

**Remark 6.** *In Eq.(22), we used the scaling of $\ln N$ to define $K$. This is motivated by the median trick of the existing PVI, which tunes the bandwidth as $\ln N/\text{median}^2$. This scaling implies that, for each $i$, $\sum_j K_{ij} \approx 1 + \frac{1}{N}$ holds. We found that using this scaling is necessary to obtain the bound Eq.(21). We conjecture that this is the reason why scaling the bandwidth is important for PVI in practice.*

In the same way as Eq.(19) for w-SGLD, we also optimize the middle term in Eq.(21) by GD. By taking the partial derivative, we have the following update equation:

$$\partial_{\theta_i} \mathbb{E}_{\rho_E(\theta)} \log p(x|\theta_i) + \frac{2}{\tilde{h}N} \sum_j (K + \epsilon I)^{-1}_{ij} \nabla_{L(\theta_j)} K_{ij} \partial_{\theta_i} L(\theta_i). \qquad (23)$$

See appendix B.7 for the proof. The second term is the repulsion force, which is the same as that in the update equation of GFSF in Table 1.

Next we consider the relation to DPPs. Using the trace inequality [8] to Eq.(16) and using a Gram matrix $\tilde{G}$ whose $(i,j)$ element is $G_{ij}^{\frac{1}{2}}$ in Eq.(17), we obtain

$$\begin{aligned}
\ln \mathbb{E}_{\rho_E} p(x|\theta) &\geq \mathbb{E}_{\rho_E} \ln p(x|\theta) + \frac{1}{N} \ln \det(I - \frac{\tilde{G}}{N}) \\
&\geq \mathbb{E}_{\rho_E} \ln p(x|\theta) \\
&\geq \mathbb{E}_{\rho_E} \ln p(x|\theta) + \frac{2}{N} \ln \det \tilde{G} - \ln N. \qquad (24)
\end{aligned}$$

The proof is shown in Appendix B.8. The lower bound term of Eq.(24) is equivalent to the objective function of the DPP introduced in Eq.(5).

For Eqs.(21) and (24), we can derive a PAC-Bayesian generalization error bound like Theorem 4 (see Appendix C.2 for details). Moreover, we can connect our loss repulsion in Eqs.(21) and (24) to model and parameter repulsions in the same way as w-SGLD. Accordingly, GFSF and DPP are closely related to the second-order generalization error bound.

In conclusion, we derived PVI and DPPs from our theory based on the second-order Jensen inequality. On the other hand, we found it difficult to show its relation to SVGD, since it has the kernel smoothing coefficient in the derivative of the log-likelihood function. We leave it for future work.

## 4 Discussion and Related work

In this section, we discuss the relationship between our theory and existing work.

### 4.1 Theoretical analysis of PVI

Originally, PVI was derived as an approximation of the gradient flow in Wasserstein space and its theoretical analysis has only been done with an infinite ensemble size [13, 11]. In practice, however, an infinite ensemble size is not realistic and various numerical experiments showed that PVI still works well even with a finite ensemble size [14, 24]. Accordingly, our analysis aimed to clarify why PVI works well with such a finite-size ensembles. On the other hand, as discussed in Section 3.2.2, there is a difference between our loss repulsion and the parameter and model repulsion used in the existing works, and thus it would be an interesting direction to extend our theory to fill such a gap.

## 4.2 Relation to existing second-order Jensen inequality

Recently, some works derived tighter Jensen inequalities [12, 6]. Liao and Berg [12] derived a second-order Jensen inequality, and Gao et al. [6] worked on a higher-order inequality. Masegosa [16] combined the earlier result [12] with the PAC-Bayesian theory. Although our bound is also a second-order Jensen inequality, its derivation and behavior are completely different from them. In Liao [12], their second-order Jensen inequality includes the term of a variance of a random variable, and Masagosa [16] considered $p(x|\theta)$ to be a corresponding random variable that depends on $\theta$. Thus, the second-order inequalities depend on the variance of the predictive distribution $\mathbb{E}_{p(\theta)}[p(x|\theta)]$. On the other hand, our bound is based on our novel second-order equality shown in Theorem 2, which leads to the variance of a loss function as shown in Theorem 3. By using the variance of loss functions, we can directly connect our theory to existing PVI and DPPs as shown in Section 3.2. Moreover, as shown in Section 5, including the predictive variance in the objective function results in a large epistemic uncertainty, which means that individual models do not fit well. On the other hand, ours does not show this phenomenon. Consequently, our result can be regarded as an extension of the earlier work [12, 6, 16] that directly focuses on a loss function in machine learning.

Masagosa [16] showed that the second-order PAC-Bayesian generalization error is especially useful under misspecified models, i.e., for any $\theta$, $p(x|\theta) \neq \nu(x)$. Our theories can also be extended to such a setting (see Appendix F for further discussion).

Other closely related work is an analysis of the weighted majority vote in multiclass classification [17], which uses a second-order PAC-Bayesian bound. While their analysis is specific to the majority vote of multiclass classification, our analysis has been carried out in a more general setting based on the second-order Jensen inequality derived from Theorem 2 and Theorem 3.

## 5 Numerical experiments

According to our Theorem 4, it is important to control the trade-off between the model fitting and diversity enhancement in order to reduce the generalization error. Therefore, we minimize our generalization error bound Eq.(15) directly and confirm that the trade-off is controlled appropriately. Our objective function is

$$\mathcal{F}(\{\theta_i\}_{i=1}^N) := -\frac{1}{N}\sum_{i=1}^N \sum_{d=1}^D [\ln p(x_d|\theta_i) + R(x_d, h)] + \mathrm{KL}(\rho_\mathrm{E}, \pi), \qquad (25)$$

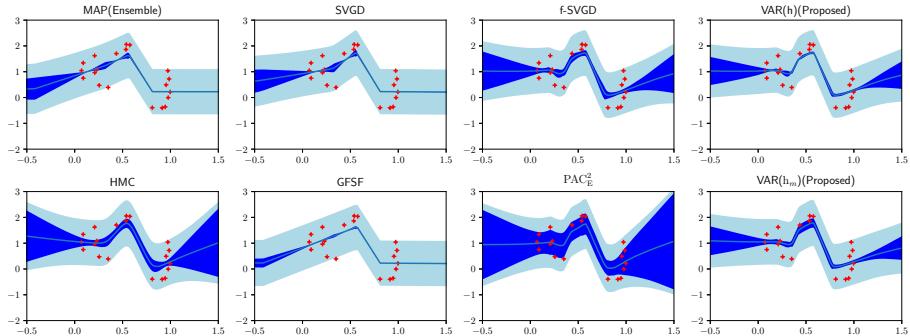

Figure 1: Uncertainty of the regressions. Blue line is the predictive mean, light shaded area and dark shaded area visualize 95% credible intervals for the prediction and mean estimate respectively.

which is obtained setting $q(\theta) = \rho_E(\theta)$ and $c = 1$ in Eq.(15). We call our approach as Variance regularization, and express it as VAR. We used two types of $h$, one is defined in Eq.(13) and the other is $h_m$ in Theorem 4. We call these approach as VAR($h$) and VAR($h_m$). We compared the performances of our methods with MAP, $\text{PAC}_E^2$, and f-PVI(f-SVGD) on toy data and real data. Our experimental settings are almost same as that of the previous work [24] and the detailed settings and additional results are shown in Appendix G.

### 5.1 Toy data experiments

First, using toy data, we visualized the model fitting and diversity in various methods. We considered a regression task and we randomly generated the 1-dimensional data points and fit them by using a feed-forward neural network model with 2 hidden layers and ReLu activation, which has 50 units. We used 50 ensembles for each method. The results are shown in Figure 1, which visualizes 95% credible intervals for the prediction and mean estimate corresponding to aleatoric and epistemic uncertainties.

Hamilton Monte Carlo (HMC)[18] is the baseline method used to express the Bayesian predictive uncertainty properly. MAP and SVGD methods give an evaluation of the uncertainty that is too small. Our method and f-SVGD showed very similar evaluations for uncertainty. $\text{PAC}_E^2$ [16] showed large epistemic uncertainty, which is expressed as dark shaded area in Figure 1. We conjecture it is because $\text{PAC}_E^2$ includes the predictive variance in the objective function and the enhanced diversity is too large. Described below, $\text{PAC}_E^2$ shows slightly worse results in real data experiments than those of other approaches. This might be because including the predictive variance in the objective function does not result in a better trade-off of between model fitting and enhancing diversity in practice.

### 5.2 Regression task on UCI

We did regression tasks on the UCI dataset [4]. The model is a single-layer network with ReLu activation and 50 hidden units except for Protein data, which has 100 units. We used 20 ensembles. Results of 20 repetition are shown in Table 2. We found that our method compares favorably with f-SVGD. We also found that PAC$_E^2$ shows worse performance than those of other methods. We conjectured that this is because the predictive variance in the objective function enhances too large diversity as shown in Figure 1, which hampers the fitting of each model to the data.

Table 2: Benchmark results on test RMSE and negative log likelihood for the regression task.

| Dataset | Avg. Test RMSE | | | | | Avg. Test negative log likelihood | | | | |
|---|---|---|---|---|---|---|---|---|---|---|
| | MAP | PAC$_E^2$ | f-SVGD | VAR(h) | VAR(h$_m$) | MAP | PAC$_E^2$ | f-SVGD | VAR(h) | VAR(h$_m$) |
| Concrete | 5.19±0.3 | 5.49±0.3 | 4.32±0.1 | 4.33±0.1 | 4.36±0.2 | 3.11±0.12 | 3.16±0.10 | 2.86±0.02 | 2.82±0.09 | 2.87±0.09 |
| Boston | 2.98±0.4 | 4.03±0.5 | 2.54±0.3 | 2.54±0.3 | 2.52±0.3 | 2.62±0.2 | 2.61±0.3 | 2.46±0.1 | 2.39±0.2 | 2.48±0.4 |
| Wine | 0.65±0.04 | 1.03±0.09 | 0.61±0.03 | 0.61±0.03 | 0.61±0.03 | 0.97±0.07 | 1.26±0.03 | 0.90±0.05 | 0.89±0.04 | 0.89±0.04 |
| Power | 3.94±0.03 | 5.04±0.21 | 3.77±0.03 | 3.76±0.03 | 3.76±0.06 | 2.79±0.05 | 3.17±0.01 | 2.76±0.05 | 2.79±0.03 | 2.76±0.02 |
| Yacht | 0.86±0.05 | 0.70±0.21 | 0.59±0.09 | 0.59±0.09 | 0.59±0.09 | 1.23±0.05 | 0.80±0.4 | 0.96±0.3 | 0.87±0.3 | 1.03±0.3 |
| Protein | 4.25±0.07 | 4.17±0.05 | 3.98±0.03 | 3.95±0.05 | 3.96±0.06 | 2.95±0.00 | 2.84±0.01 | 2.80±0.01 | 2.81±0.01 | 2.80±0.01 |

Table 3: Benchmark results on test accuracy and negative log likelihood for the classification task.

| Dataset | Test Accuracy | | | | | Test log likelihood | | | | |
|---|---|---|---|---|---|---|---|---|---|---|
| | MAP | PAC$_E^2$ | f-SVGD | VAR(h) | VAR(h$_m$) | MAP | PAC$_E^2$ | f-SVGD | VAR(h) | VAR(h$_m$) |
| Mnist | 0.981 | 0.986 | 0.987 | 0.988 | 0.988 | 0.057 | 0.042 | 0.043 | 0.040 | 0.041 |
| Cifar 10 | 0.935 | 0.919 | 0.927 | 0.928 | 0.927 | 0.215 | 0.270 | 0.241 | 0.238 | 0.242 |

Table 4: Cumulative regret relative to that of the uniform sampling.

| Dataset | MAP | PAC$_E^2$ | f-SVGD | VAR(h) | VAR(h$_m$) |
|---|---|---|---|---|---|
| Mushroom | 0.129±0.098 | 0.037±0.012 | 0.043±0.009 | **0.029±0.010** | 0.036±0.012 |
| Financial | 0.791±0.219 | 0.189±0.025 | 0.154±0.017 | 0.155±0.024 | **0.128±0.017** |
| Statlog | 0.675±0.287 | 0.032±0.0025 | 0.010±0.0003 | **0.006±0.0003** | 0.008±0.0005 |
| CoverType | 0.610±0.051 | 0.396±0.006 | 0.372±0.007 | **0.289±0.003** | 0.343±0.002 |

### 5.3 Classification task on MNIST and CIFAR 10

We conducted numerical experiments on MNIST and CIFAR 10 datasets. For MNIST, we used a feed-forward network having two hidden layers with 400 units and a ReLu activation function and used 10 ensembles. For CIFAR 10, we used ResNet-32 [9], and we used 6 ensembles. The results are shown in Table 3. For both datasets, our methods show competitive performance compared to f-SVGD. For CIFAR 10, as reported previously [24], f-SVGD is worse than the simple ensemble approach. We also evaluated the diversity enhancing property by using the out of distribution performance test. It is hypothesized that Bayesian models are more robust against adversarial examples due to their ability to capture uncertainty. Thus, we generated attack samples and measured the vulnerability to those samples. We found that, as shown in Figure 4 in Appendix G, our method and f-SVGD showed more robustness compared to MAP estimation in each experiment.

### 5.4 Contextual bandit by neural networks on the real data set

Finally, we evaluated the uncertainty of the obtained models using contextual bandit problems [21]. This problem requires the algorithm to balance the trade-off between the exploitation and exploration, and poorly evaluated uncertainty results in larger cumulative regret. We consider the Thompson sampling algorithm with Bayesian neural networks having 2 hidden layers and 100 ReLu units, and we used 20 particles for each experiments. Results of 10 repetition are shown in Table 4. We can see that our approach outperform other methods.

# 6 Conclusion

In this work, we derived a novel second-order Jensen inequality that includes the variance of loss functions. We also derived a PAC-Bayesian generalization error bound. Our error bound shows that both model fitting and enhancing diversity are important for reducing the generalization error. Then, we derived the existing PVI and DPPs from our new Jensen inequality. Finally, we proposed a new PVI that directly minimizes our PAC-Bayesian bound. It shows competitive performance with the current state-of-the-art PVI. In future work, it would be interesting to apply our second-order Jensen inequality to general variational inference or optimal control problems.

Other interesting direction is to derive an upper-bound of the Jensen gap. Gao et al. [6] derived $\ln \mathbb{E} p(x|\theta) - \mathbb{E} \ln p(x|\theta) \leq \text{Variacne}(p(x|\theta))$, which uses the predictive distribution. This means that the larger the predictive variance is, the larger upper bound of the Jensen gap we have. We leave it for future work to upper-bound the Jensen gap with the variance of loss function using our theory.

# Acknowledgements

FF was supported by JST ACT-X Grant Number JPMJAX190R.

# A  Further preliminary of existing methods

In this section, we review existing methods.

## A.1  Function space PVI

Here we review the Function space PVI (f-PVI) [24]. Pairs of input-output data are expressed as $\mathcal{D} = \{(x,y)\}$ where $x \in \mathcal{X}$ and $y \in \mathcal{Y}$. We consider the model $p(y|x,\theta) = p(y|f(x;\theta))$ where $f(x;\theta)$ is a $c$-dimensional output function parametrized by $\theta$ and $x$ is an input. For example, for regression tasks, if $\mathcal{Y} = \mathbb{R}$, then $c = 1$ and we often assume that $p(y|x,\theta) = N(y|f(x;\theta), \sigma^2)$. For classification tasks, $c$ corresponds to the class number and $p(y|x,\theta) = \text{Multinomial}(y|\text{softmax}(f(x;\theta)))$.

In previous work [24], they considered that there exists a mapping from the parameters $\theta$ to a function $f(\cdot, \theta)$, and a prior distribution on $\theta$ implicitly defines a prior distribution on the space of the function, $\pi(f)$. Then, the model $p(y|x,\theta)$ corresponds to the distribution of $p(y|x,f)$. Thus, it is possible to obtain the posterior distribution for function $f$ from the inference of the parameters.

Then, we approximate the distribution on $f$ by a size-N ensemble set $\{f_i(\cdot)\}_{i=1}^N$. That is, we directly update each $f_i(\cdot)$. Then, the update of f-PVI is given as

$$f_{\text{new}}^i(\mathcal{X}) \leftarrow f_{\text{old}}^i(\mathcal{X}) + \eta v(\{f_i(\mathcal{X})\}_{i=1}^N), \tag{26}$$

where $v$ is the update direction.

However, if the input space $\mathcal{X}$ is very large or infinite, it is impossible to directly update $f_i(\cdot)$ efficiently. Instead, in the previous work [24], we approximate $f_i(\cdot)$ by a parametrized neural network. In principle, it is possible to use any flexible network to approximate it, and it was proposed to use the original network, $f(\cdot;\theta)$ for that approximation since it can express any function on the support of the prior $\pi(f)$ which is implicitly defined by the prior $\pi(\theta)$. Thus, the update direction is mapped to the parameter space

$$\theta_i^{\text{new}} \leftarrow \theta_i^{\text{old}} + \eta \frac{\partial f_i(\mathcal{X})}{\partial \theta_i}\Big|_{\theta_i = \theta_i^{\text{old}}} v(\{f_i(\mathcal{X})\}_{i=1}^N). \tag{27}$$

Furthermore, it was proposed to replace $\mathcal{X}$ in Eq.(27) by a finite set of samples $\boldsymbol{x}_{1:b} = (x_1, \ldots, x_b)$ with size $b$, which are drawn from $\mathcal{X}^{\otimes b}$. When we input the minibatch with size $b$ into the model, we express it as $f_i(\boldsymbol{x}_{1:b}) = (f_i(x_1), \ldots, f_i(x_b)) \in \mathbb{R}^{cb}$. Then the update equation is given as

$$\theta_i^{\text{new}} \leftarrow \theta_i^{\text{old}} + \eta \frac{\partial f_i(\boldsymbol{x}_{1:b})}{\partial \theta_i}\Big|_{\theta_i = \theta_i^{\text{old}}} v(\{f_i(\boldsymbol{x}_{1:b})\}_{i=1}^N), \tag{28}$$

where $v(\{f_i\})$ is obtained by replacing $\ln p(\mathcal{D}|\theta)\pi(\theta)$ with $(D/b)\sum_{d=1}^b \ln p(x_d|\theta)\pi(f)$, where $\pi(f)$ is a prior distribution over $f$ and the Gram matrix $K(\theta_i, \theta_j)$ is replaced with $K(f_i^b, f_j^b)$ in Table 1. Thus, f-PVI modifies the loss signal so that models are diverse. We refer to the repulsion term of f-PVI as a model repulsion. When we use the Gaussian kernel, the model repulsion is expressed as

$$\partial_{\theta_i} K(f_i^b, f_j^b) = -h^{-2}(f_i^b - f_j^b) e^{-\|f_i^b - f_j^b\|^2/(2h^2)} \partial_{\theta_i} f_i^b. \tag{29}$$

Thus, the model repulsion pushes model $f_i$ away from $f_j$.

Finally, the implicitly defined prior $\pi(f)$ should be specified more explicitly to calculate $\partial_f \ln \pi(f)$. Then it was proposed approximating the implicit prior by a Gaussian process. Given a minibatch of data $\boldsymbol{x}$, first, we draw parameters from a prior $\pi(\theta)$ and then construct a multivariate Gaussian distribution whose mean and variance are defined by the mean and variance of drawn samples.

## A.2 The second-order PAC-Bayesian generalization error bound [16]

We first present the second-order PAC-Baysian bound in previous work [16],

**Theorem 7.** *[16] For all $x, \theta$, $p(x|\theta) < \infty$ and for any prior distribution $\pi$ over $\Theta$ independent of $\mathcal{D}$ and for any $\xi \in (0, 1)$ and $c > 0$, with probability at least $1 - \xi$ over the choice of training data $\mathcal{D} \sim \nu^{\otimes D}(x)$, for all probability distributions $q$ over $\Theta$, we have*

$$
\begin{aligned}
\mathrm{CE} &\leq -\mathbb{E}_{\nu,q}[\ln p(x|\theta) + V(x)] \\
&\leq \mathbb{E}_q \frac{-1}{D} \sum_{d=1}^{D} [\ln p(x_d|\theta) + V(x_d)] + \frac{\mathrm{KL}(q, \pi) + \frac{\ln \xi^{-1} + \Psi'_{\pi,\nu}(c,D)}{2}}{cD},
\end{aligned} \quad (30)
$$

where

$$
V(x) := (2 \max_\theta p(x|\theta)^2)^{-1} \mathbb{E}_{q(\theta)}\left[(p(x|\theta) - \mathbb{E}_{q(\theta)} p(x|\theta))^2\right], \quad (31)
$$

and

$$
\Psi''_{\pi,\nu}(c, D) := \ln \mathbb{E}_{\pi(\theta,\theta')} \mathbb{E}_{\mathcal{D}\sim\nu^{\otimes D}(x)} e^{cD(-\mathbb{E}_{\nu(x)} L(x,\theta,\theta') + D^{-1} \sum_{d=1}^{D} L(x_d,\theta,\theta'))}, \quad (32)
$$

and

$$
L(x, \theta, \theta') := \ln p(x|\theta) + (2 \max_\theta p(x|\theta)^2)^{-1} (p(x|\theta)^2 - p(x|\theta) p(x|\theta')). \quad (33)
$$

## A.3 Ensemble setting of the second-order PAC-Bayesian generalization error bound [16]

Here we introduce the ensemble setting, that is,

$$
\rho_{\mathrm{E}}(\theta) = \frac{1}{N} \sum_{i=1}^{N} \delta_{\theta_i}(\theta). \quad (34)
$$

### A.3.1 Prior distribution for ensemble learning

We need to properly define a prior distribution for ensemble learning so that KL divergence between $\rho_{\mathrm{E}}$ and $\pi$ can be defined properly. Following the previous work [16], we define the prior distribution as the mixture of discrete dirac mass distribution as

$$
\pi_{\mathrm{E}}(\theta) = \sum_{\theta' \in \Theta_{\mathrm{E}}} w_{\theta'} \delta_{\theta'}(\theta), \quad (35)
$$

where $w_{\theta'} \geq 0$ and $\sum_{\theta' \in \Theta_{\mathrm{E}}} w_{\theta'} = 1$.
If

$$
\{\theta_i\}_{i=1}^{N} \subset \Theta_{\mathrm{E}} \subset \mathbb{R}^d, \quad (36)
$$

holds, we can define the KL divergence properly. From the definition of Radon–Nikodym derivative of the discrete measure, we have

$$\mathrm{KL}(\rho_\mathrm{E}, \pi_\mathrm{E}) = \frac{1}{N}\sum_{i=1}^{N} \ln \frac{\frac{1}{N}}{w_{\theta_i}} = -\frac{1}{N}\sum_{i=1}^{N} \ln \pi_\mathrm{E}(\theta_i) + \frac{1}{N}\sum_{i=1}^{N} \ln \frac{1}{N}. \tag{37}$$

Thus, we need to properly define $\Theta_\mathrm{E}$ so that Eq.(34) holds.

Following the idea of [16], we define $\Theta_\mathrm{E}$ as the set of $d$-dimensional real vectors that can be represented under a finite-precision scheme using $p$-bits to encode each element of the vector. Thus this set is countable and can define the mixture of dirac mass distribution and satisfies Eq.(34) properly.

Note that as discussed in the previous work [16], the KL divergence of Eq.(37) is not continuous and differentiable and not suitable for the gradient descent based optimization. Fortunately, when we implement any statistical distribution on a computer, they are expressed under a finite-precision scheme, thus, we can regard them as an approximation of $\pi_\mathrm{E}$. Thus, we can use any statistical distribution as a proxy of a precise $\pi_\mathrm{E}$ when we implement algorithms on a computer.

### A.3.2 Generalization error bound

Using the prior distribution $\pi_\mathrm{E}$ introduced in Appendix A.3.1, we have the PAC-Bayesian generalization error bound for the ensemble setting,

**Theorem 8.** *[16] For all $x, \theta$, $p(x|\theta) < \infty$ and for any prior distribution $\pi_\mathrm{E}$ over $\Theta_\mathrm{E}$ independent of $\mathcal{D}$ and for any $\xi \in (0,1)$ and $c > 0$, with probability at least $1 - \xi$ over the choice of training data $\mathcal{D} \sim \nu^{\otimes D}(x)$, for all probability distributions $\rho_\mathrm{E}$ with $\mathrm{supp}(\rho_\mathrm{E}) \subset \Theta_\mathrm{E}$, we have*

$$\begin{aligned}\mathrm{CE} &\leq -\mathbb{E}_{\nu,\rho_\mathrm{E}}[\ln p(x|\theta) + V(x)] \\ &\leq \mathbb{E}_{\rho_\mathrm{E}} \frac{-1}{D}\sum_{d=1}^{D}[\ln p(x_d|\theta) + V(x_d)] + \frac{\mathrm{KL}(\rho_\mathrm{E}, \pi_\mathrm{E}) + \frac{\ln \xi^{-1} + \Psi'_{\pi_\mathrm{E},\nu}(c,D)}{2}}{cD},\end{aligned} \tag{38}$$

where

$$V(x) := (2\max_{\theta} p(x|\theta)^2)^{-1}\mathbb{E}_{\rho_\mathrm{E}}\left[(p(x|\theta) - \mathbb{E}_{\rho_\mathrm{E}} p(x|\theta))^2\right], \tag{39}$$

and

$$\Psi''_{\pi_\mathrm{E},\nu}(c,D) := \ln \mathbb{E}_{\pi(\theta,\theta')} \mathbb{E}_{\mathcal{D}\sim\nu^{\otimes D}(x)} \mathrm{e}^{cD(-\mathbb{E}_{\nu(x)} L(x,\theta,\theta') + D^{-1}\sum_{d=1}^{D} L(x_d,\theta,\theta'))}, \tag{40}$$

and

$$L(x,\theta,\theta') := \ln p(x|\theta) + (2\max_{\theta} p(x|\theta)^2)^{-1}(p(x|\theta)^2 - p(x|\theta)p(x|\theta')). \tag{41}$$

# B Proofs of Section 3

## B.1 Proof of Theorem 2

Since $\psi$ is a monotonically increasing concave function on $\mathbb{R}$, an inverse function $\psi^{-1}$ is convex function.

Let us consider the Taylor expansion of $\psi$ up to the second order around a constant $\mu$. There exists a constant $c(z)$ between $\mu$ and $z$ that satisfies

$$\psi(z) = \psi(\mu) + \psi'(\mu)(z-\mu) + \frac{\psi''(c(z))}{2}(z-\mu)^2. \tag{42}$$

Above equation holds for all realizable values of a random variable $Z$.

Then, we substitute a random variable $Z$ to $z$ and $\mu = \psi^{-1}(\mathbb{E}_{p_Z}[\psi(Z)])$ in the above equation,

$$0 = \psi'(\mu)\int_{\mathbb{R}^+}(z-\mu)p_Z(z)dz + \int_{\mathbb{R}^+}\frac{\psi''(c(z))}{2}(z-\mu)^2 p_Z(z)dz. \quad (43)$$

By rearranging the above, and $\psi' > 0$ since $\psi'$ is a monotonically increasing function, we have

$$\mathbb{E}_{p_Z}Z = \mu - \int_{\mathbb{R}^+}\frac{\psi''(c(z))}{2\psi'(\mu)}(z-\mu)^2 p_Z(z)dz \quad (44)$$

Since $\mathbb{E}_{p_Z}Z \in \mathbb{R}^+$ from the assumption and since $\psi$ is a monotonically increasing function, $\psi'$ is always positive and $\psi''$ is always negative. Thus $\mu - \mathbb{E}_{p_Z}\frac{\psi''(c)}{2\psi'(\mu)}(Z-\mu)^2 \in \mathbb{R}^+$. Then we apply the $\psi$ on both hand side, we obtain the theorem.

## B.2 Proof of Corollary 1

**Remark 7.** *This corollary holds for all probability distributions $q$ over $\Theta$, like Theorem 1.*

*Proof.* We substitute $\psi = \log$ and $Z = p(x|\theta)$ in Theorem 2, we obtain the result. Here we also show the more intuitive proof.

Let us consider the Taylor expansion of log function up to the second order around a constant $\mu = ^{\mathbb{E}_q \ln p(x|\theta)}$. We obtain

$$\ln e^{\ln p(x|\theta)} = \ln e^{\mathbb{E}_q \ln p(x|\theta)}$$
$$+ \frac{1}{e^{\mathbb{E}_q \ln p(x|\theta)}}(e^{\ln p(x|\theta)} - e^{\mathbb{E}_q \ln p(x|\theta)}) - \frac{1}{2g(x,\theta)^2}(e^{\ln p(x|\theta)} - e^{\mathbb{E}_q \ln p(x|\theta)})^2 \quad (45)$$

where $g(x, \theta)$ is the constant between $p(x|\theta)$ and $\mu$, and it is defined as the reminder of the Taylor expansion. Taking the expectation, we have

$$0 = \mathbb{E}_q \frac{1}{e^{\mathbb{E}_q \ln p(x|\theta)}}(e^{\ln p(x|\theta)} - e^{\mathbb{E}_q \ln p(x|\theta)}) - \mathbb{E}_q \frac{1}{2g(x,\theta)^2}(e^{\ln p(x|\theta)} - e^{\mathbb{E}_q \ln p(x|\theta)})^2 \quad (46)$$

We rearrange the equality as follows

$$e^{\mathbb{E}_q \ln p(x|\theta)}\left(1 + \mathbb{E}_q \frac{1}{2g(x,\theta)^2}(e^{\ln p(x|\theta)} - e^{\mathbb{E}_q \ln p(x|\theta)})^2\right) = \mathbb{E}_q e^{\ln p(x|\theta)}. \quad (47)$$

Then taking the logarithm in both hand side, we obtain the result. $\square$

## B.3 Proof of Theorem 3

**Remark 8.** *This theorem holds for all probability distributions $q$ over $\Theta$, like Theorem 1.*

*Proof.* First, recall that $g(x, \theta)$ is the constant between $p(x|\theta)$ and $\mu$ is defined as the reminder of the second order Taylor expansion. Thus if we define $g(x) := \max_{\theta \in \mathrm{supp}(q(\theta))} p(x|\theta)$, then $g(x) \geq g(x, \theta)$ holds. Moreover, following relation holds:

$$\frac{1}{2}\left(\frac{e^{\ln p(x|\theta)} - e^{\mathbb{E}_q \ln p(x|\theta)}}{g(x)}\right)^2 \leq 1. \quad (48)$$

Using the following lemma (its proof is shown in the below)

**Lemma 1.** *For any constant $\alpha \in (0, 1]$, we have*

$$-\ln(1+\alpha) \leq \ln(1 - \frac{\alpha}{2}). \tag{49}$$

Then we obtain

$$\mathbb{E}_q \ln p(x|\theta) \leq \ln \mathbb{E}_q p(x|\theta) + \ln\left(1 - \mathbb{E}_q \frac{1}{4g(x)^2}(e^{\ln p(x|\theta)} - e^{\mathbb{E}_q \ln p(x|\theta)})^2\right). \tag{50}$$

Then using the relation

$$1 - \frac{\alpha}{2} \leq e^{-\frac{\alpha}{2}}, \tag{51}$$

we get

$$\mathbb{E}_q \ln p(x|\theta) \leq \ln \mathbb{E}_q p(x|\theta) + \ln e^{-\mathbb{E}_q \left(\frac{e^{\ln p(x|\theta)} - e^{\mathbb{E}_q \ln p(x|\theta)}}{2g}\right)^2}. \tag{52}$$

Finally, we use the following lemma (the proof is shown in the below),

**Lemma 2.** *For any positive constants $\alpha, \beta > 0$, we have*

$$\sqrt{\alpha\beta} \leq \frac{\alpha - \beta}{\ln \alpha - \ln \beta}, \tag{53}$$

*and when $\alpha = \beta$, the equality holds.*

Above lemma is equivalent to

$$(\ln \alpha - \ln \beta)^2 \alpha\beta \leq (\alpha - \beta)^2. \tag{54}$$

Setting $\alpha := e^{\ln p(x|\theta)}$ and $\beta := e^{\mathbb{E}_q \ln p(x|\theta)}$ and substituting them into Eq.(54), we obtain

$$(\ln p(x|\theta) - \mathbb{E}_q \ln p(x|\theta))^2 p(x|\theta) e^{\mathbb{E}_q \ln p(x|\theta)} \leq (e^{\ln p(x|\theta)} - e^{\mathbb{E}_q \ln p(x|\theta)})^2. \tag{55}$$

Then we obtain

$$\mathbb{E}_q \ln p(x|\theta) \leq \ln \mathbb{E}_q p(x|\theta) + \ln e^{-\mathbb{E}_q p(x|\theta) e^{\mathbb{E}_q \ln p(x|\theta)} \left(\frac{(\ln p(x|\theta) - \mathbb{E}_q \ln p(x|\theta))}{2g(x)}\right)^2}. \tag{56}$$

We define the bandwidth as

$$p(x|\theta) e^{\mathbb{E}_q \ln p(x|\theta)} / g(x)^2 := h(x, \theta)^{-2}, \tag{57}$$

then, we have

$$\mathbb{E}_q \ln p(x|\theta) \leq \ln \mathbb{E}_q p(x|\theta) + \ln e^{-\mathbb{E}_q \left(\frac{\ln p(x|\theta) - \mathbb{E}_q \ln p(x|\theta)}{2h(x,\theta)}\right)^2}. \tag{58}$$

□

### B.3.1 Proof of lemma 1

*Proof.* Define $f(\alpha) := \ln(1 - \frac{\alpha}{2}) + \ln(1 + \alpha)$. Since $f'(\alpha) > 0$ for $0 \leq \alpha < 1$, thus $0 = f(0) \leq f(\alpha)$. This concludes the proof. □

### B.3.2 Proof of lemma 2

*Proof.* Since

$$\frac{\alpha - \beta}{\ln \alpha - \ln \beta} \geq 0, \tag{59}$$

we only need to show

$$\frac{\ln \alpha - \ln \beta}{\alpha - \beta} \leq \frac{1}{\sqrt{\alpha \beta}}. \tag{60}$$

Since this inequality is symmetric with respect to $\alpha$ and $\beta$, we can assume that $\alpha \geq \beta$. If $\alpha = \beta$, it is clear by setting $\beta = \alpha + \epsilon$ where $\epsilon > 0$ and take the limit to $\epsilon \to 0^+$. When $\alpha > \beta$, we define $\alpha = t\beta$ where $t > 1$. Substituting this assumption in the above, we need to show that

$$\frac{2 \ln t}{t^2 - 1} \leq \frac{1}{t}. \tag{61}$$

Since $t^2 - 1 > 0$, by rearranging the above inequality, we only need to show that

$$t - \frac{1}{t} - 2 \ln t \geq 0. \tag{62}$$

Thus, we define $f(t) := t - \frac{1}{t} - 2 \ln t$. Then $f'(t) \geq 0$ for all $t$, thus $0 = f(1) \leq f(t)$ for all $t \geq 1$. This concludes the proof. □

## B.4 Proof of Theorem 4

We first present the complete statement:

**Theorem 9.** *For all $x$ and $\theta$, $p(x|\theta) < \infty$ and for any prior distribution $\pi$ over $\Theta$ independent of $\mathcal{D}$ and for any $\xi \in (0,1)$ and $c > 0$, with probability at least $1 - \xi$ over the choice of training data $\mathcal{D} \sim \nu^{\otimes D}(x)$, for all probability distribution $q$ over $\Theta$, we have*

$$\begin{aligned} \mathrm{CE} &\leq -\mathbb{E}_{\nu,q}[\ln p(x|\theta) + R(x,h)] \\ &\leq \mathbb{E}_q \frac{-1}{D} \sum_{d=1}^{D} [\ln p(x_d|\theta) + R(x_d, h_m)] + \frac{\mathrm{KL}(q,\pi) + \frac{\ln \xi^{-1} + \Psi''_{\pi,\nu}(c,D)}{3}}{cD}, \end{aligned} \tag{63}$$

*where*

$$\Psi''_{\pi,\nu}(c, D) := \ln \mathbb{E}_{\pi(\theta,\theta',\theta'')} \mathbb{E}_{\mathcal{D} \sim \nu^{\otimes D}(x)} e^{cD(-\mathbb{E}_{\nu(x)} L(x,\theta,\theta',\theta'') + D^{-1} \sum_{d=1}^{D} L(x_d,\theta,\theta',\theta''))}, \tag{64}$$

*and*

$$\begin{aligned} &L(x, \theta, \theta', \theta'') \\ &:= \ln p(x|\theta) + (2h_m(x,\theta))^{-2}((\ln p(x|\theta))^2 - 2\ln p(x|\theta) \ln p(x|\theta') + \ln p(x|\theta') \ln p(x|\theta'')). \end{aligned} \tag{65}$$

*Proof.* From the definition of the band width, we have

$$\mathrm{CE} \leq -\mathbb{E}_{\nu,q}[\ln p(x|\theta) + R(x,h)] \leq -\mathbb{E}_{\nu,q}[\ln p(x|\theta) + R(x,h_m)]. \tag{66}$$

Thus, our goal is to derive the probabilistic relationship

$$-\mathbb{E}_{\nu,q}[\ln p(x|\theta) + R(x,h_m)] \approx \mathbb{E}_q \frac{-1}{D} \sum_{d=1}^{D} [\ln p(x_d|\theta) + R(x_d, h_m)]. \tag{67}$$

We express $\mathbb{E}_{q(\theta)}[\ln p(x|\theta) + R(x, h_m)]$ as

$$\mathbb{E}_{q(\theta)q(\theta')q(\theta'')} L(x, \theta, \theta', \theta''), \tag{68}$$

where

$$L(x, \theta, \theta', \theta'')$$
$$:= \ln p(x|\theta) + (2h_m(x,\theta))^{-2}((\ln p(x|\theta))^2 - 2\ln p(x|\theta) \ln p(x|\theta') + \ln p(x|\theta') \ln p(x|\theta'')). \tag{69}$$

Then, we consider the same proof as Pac-Bayesian bound of Theorem 1 [7] to this loss function. Applying Theorem 3 in [7] with a prior $\pi(\theta, \theta', \theta'') := \pi(\theta)\pi(\theta')\pi(\theta'')$, we have

$$\begin{aligned}
&\mathbb{E}_{\nu(x),q(\theta)q(\theta')q(\theta'')} L(x, \theta, \theta', (\theta'') \\
&\leq -\frac{1}{D} \sum_d^D \mathbb{E}_{q(\theta)q(\theta')q(\theta'')} L(x_d, \theta, \theta', \theta'') \\
&+ \frac{\mathrm{KL}(q(\theta)q(\theta')q(\theta'')|\pi(\theta, \theta', \theta'')) + \ln \xi^{-1} + \Psi''_{\pi,\nu}(\lambda, D)}{\lambda},
\end{aligned} \tag{70}$$

where

$$\Psi''_{\pi,\nu}(c, D)$$
$$:= \ln \mathbb{E}_{\pi(\theta,\theta',\theta'')} \mathbb{E}_{\mathcal{D} \sim \nu^{\otimes D}(x)} \mathrm{e}^{cD(-\mathbb{E}_{\nu(x)} L(x,\theta,\theta',\theta'') + D^{-1} \sum_{d=1}^D L(x_d,\theta,\theta',\theta''))}. \tag{71}$$

Noting that $\mathrm{KL}(q(\theta)q(\theta')q(\theta'')|\pi(\theta, \theta', \theta'')) = 3\mathrm{KL}(q(\theta)|\pi(\theta))$, reparametrizing $\lambda = 3cD$, we obtain the main result. □

### B.4.1 Median lower bound

The goal of this section is to show that if $\mathbb{E}_q[\ln p(x|\theta)]^2 < M < \infty$, then we have

$$\mathrm{Med}[e^{\ln p(x|\theta)}]e^{-M^{1/2}} \leq e^{\mathbb{E}_q \ln p(x|\theta)}, \tag{72}$$

where Med is the median of the random variable. We relax the condition of the bandwidth $h_m$ in Theorem 4. To derive Theorem 4, we lower-bound $e^{\mathbb{E}_q \ln p(x|\theta)}$ by $e^{\min_\theta p(x|\theta)}$ and introduced $h_m$. This $e^{\min_\theta p(x|\theta)}$ can be a small value for many practical models. If we can use Eq.(72), we can replace $e^{\min_\theta p(x|\theta)}$ in $h_m$ with $\mathrm{Med}[e^{\ln p(x|\theta)}]e^{-M^{1/2}}$, which results in a much tighter bound.

*Proof.* We use the following lemma in the previous work [10],

**Lemma 3.** *Given a random variable* $\mathbb{E}X^2 < \infty$, *we have*

$$|\mathbb{E}[X] - \mathrm{Med}[X]| \leq \sqrt{\mathbb{E}(X - \mathbb{E}X)^2}. \tag{73}$$

For the completeness, we show its proof.

*Proof.* Since the median minimizes the mean absolute value error, we have

$$\begin{aligned}
|\mathbb{E}[X] - \mathrm{Med}[X]| &= |\mathbb{E}[X - \mathrm{Med}[X]]| \\
&\leq \mathbb{E}|X - \mathrm{Med}[X]| \\
&\leq \mathbb{E}|X - \mathbb{E}X| \\
&\leq \sqrt{\mathbb{E}(X - \mathbb{E}X)^2}.
\end{aligned} \tag{74}$$

□

From this lemma, we have

$$\text{Med}[X] - \sqrt{\mathbb{E}(X - \mathbb{E}X)^2} \leq \mathbb{E}X, \tag{75}$$

and taking the exponential, we have

$$e^{\text{Med}[X]} e^{-\sqrt{\mathbb{E}(X - \mathbb{E}X)^2}} \leq e^{\mathbb{E}X}. \tag{76}$$

From the definition of the median and the monotonically increasing property of the exponential function, we have

$$e^{\text{Med}[X]} = \text{Med}[e^X]. \tag{77}$$

Thus, we have

$$\text{Med}[e^X] e^{-\sqrt{\mathbb{E}(X - \mathbb{E}X)^2}} \leq e^{\mathbb{E}X}. \tag{78}$$

Finally, by setting $X = \ln p(x|\theta)$, and assume that $\mathbb{E}_q[\ln p(x|\theta)]^2 < M < \infty$, we have

$$\text{Med}[e^{\ln p(x|\theta)}] e^{-\sqrt{\mathbb{E}_q(\ln p(x|\theta) - \mathbb{E}_q \ln p(x|\theta))^2}} \leq e^{\mathbb{E}_q \ln p(x|\theta)}, \tag{79}$$

thus, we have

$$\text{Med}[e^{\ln p(x|\theta)}] e^{-M^{1/2}} \leq e^{\mathbb{E}_q \ln p(x|\theta)}. \tag{80}$$

$\square$

## B.5 Proof of Theorem 5

**Remark 9.** *This theorem and the results of Section 3.2 holds for all probability distributions $\rho_{\text{E}}$ over $\Theta$ that is expressed as the mixture of the dirac distribution, that is, $\rho_{\text{E}}(\theta) = \frac{1}{N} \sum_{i=1}^{N} \delta_{\theta_i}(\theta)$.*

*Proof.* First note that, from Theorem 3, by substituting the definition of $\rho_{\text{E}}$, we have

$$\mathbb{E}_{\rho_{\text{E}}} \ln p(x|\theta) \leq \ln \mathbb{E}_{\rho_{\text{E}}} p(x|\theta) - \frac{1}{N} \sum_{i=1}^{N} \left( \frac{\ln p(x|\theta_i) - \frac{1}{N} \sum_{j=1}^{N} \ln p(x|\theta_j)}{2h(x,\theta)} \right)^2. \tag{81}$$

Here the bandwidth $h(x, \theta)$ is

$$h(x,\theta)^{-2} = \frac{e^{\ln p(x|\theta_i) + \frac{1}{N} \sum_{i=1} \ln p(x|\theta_i)}}{e^{2 \max_{j \in \{1,\ldots,N\}} \ln p(x|\theta_j)}}. \tag{82}$$

and we simply express the $\max_{j \in \{1,\ldots,N\}} \ln p(x|\theta_j)$ as $\max_j \ln p(x|\theta_j)$. This is because that the bandwidth $h(x, \theta)$ is derived by the relation $g(x) \geq g(x, \theta)$ in Appendix B.3 and $g(x, \theta)$ is the constant between $p(x|\theta)$ and $e^{\mathbb{E}_q \ln p(x|\theta)}$. Thus, we can upper bound $g(x, \theta)$ by $\max_j \ln p(x|\theta_j)$ since $\rho_{\text{E}}$ takes values only on $\theta_1, \ldots, \theta_N$.

For that purpose, we first eliminate the dependence of $\theta$ from the bandwidth. From the definition, Let us define

$$h_w(x,\theta)^{-2} := \frac{e^{\min_i \ln p(x|\theta_i) + \frac{1}{N} \sum_{i=1} \ln p(x|\theta_i)}}{e^{2 \max_j \ln p(x|\theta_j)}}, \tag{83}$$

then we have

$$h_w(x,\theta)^{-2} \leq h(x,\theta)^{-2}. \tag{84}$$

Thus, we have

$$\mathbb{E}_{\rho_E} \ln p(x|\theta) \leq \ln \mathbb{E}_{\rho_E} p(x|\theta) - \frac{1}{4h_w(x,\theta)^2 N} \sum_{i=1}^{N} \left( \ln p(x|\theta_i) - \frac{1}{N} \sum_{j=1}^{N} \ln p(x|\theta_j) \right)^2. \tag{85}$$

Next we focus on the following relation for the variance. For simplicity, we express $L_i := \ln p(x|\theta_i)$. Then by rearranging the definition,

$$\frac{1}{N} \sum_{i=1}^{N} (L_i - \sum_{j=1}^{N} \frac{1}{N} L_j)^2$$

$$= \frac{1}{N^3} \sum_{i=1}^{N} \left( N^2 L_i^2 - 2NL_i(\sum_{j=1}^{N} L_j) + (\sum_{j=1}^{N} L_j)^2 \right)$$

$$= \frac{1}{N^3} \sum_{i=1}^{N} \left( N^2 L_i^2 - 2NL_i(\sum_{j=1}^{N} L_j) + \sum_{j=1}^{N} L_j^2 + 2 \sum_{j=1}^{N-1} L_j L_{j+1} \right)$$

$$= \frac{1}{N^3} \sum_{i=1}^{N} \left( \sum_{j=1}^{N} (L_j - L_i)^2 + (N^2 - N)L_i^2 - 2(N-1)L_i \sum_{j=1}^{N} L_j + 2\sum_{j=1}^{N-1} L_j L_{j+1} \right)$$

$$= \frac{1}{N^3} \left( \sum_{i,j=1}^{N} (L_j - L_i)^2 + (N^2 - N) \sum_{i=1}^{N} L_i^2 - 2(N-1)(\sum_{i=1}^{N} L_i)^2 + 2N \sum_{i=1}^{N-1} L_i L_{i+1} \right)$$

$$= \frac{1}{N^3} \left( \sum_{i,j=1}^{N} (L_j - L_i)^2 + (N^2 - 3N + 2) \sum_{i=1}^{N} L_i^2 + (-2N+4) \sum_{i=1}^{N-1} L_i L_{i+1} \right)$$

$$= \frac{1}{N^3} \left( \sum_{i,j=1}^{N} (L_j - L_i)^2 + (N^2 - 2N) \sum_{i=1}^{N} L_i^2 + (-N+2) \left( \sum_{i=1}^{N} L_i \right)^2 \right)$$

$$= \frac{1}{N^3} \sum_{i,j=1}^{N} (L_j - L_i)^2 + \frac{(N-2)}{N} \left( \frac{1}{N} \sum_{i=1}^{N} (L_i)^2 - \left( \sum_{i=1}^{N} \frac{L_i}{N} \right)^2 \right)$$

$$= \frac{1}{N^3} \sum_{i,j=1}^{N} (L_j - L_i)^2 + \frac{(N-2)}{N} \left( \frac{1}{N} \sum_{i=1}^{N} (L_i - \sum_{j=1}^{N} \frac{1}{N} L_j)^2 \right). \tag{86}$$

Thus, we have

$$\frac{2}{N} \frac{1}{N} \sum_{i=1}^{N} (L_i - \sum_{j=1}^{N} \frac{1}{N} L_j)^2 = \frac{1}{N^3} \sum_{i,j=1}^{N} (L_j - L_i)^2, \tag{87}$$

which means

$$\frac{1}{N} \sum_{i=1}^{N} (L_i - \sum_{j=1}^{N} \frac{1}{N} L_j)^2 = \frac{1}{2N^2} \sum_{i,j=1}^{N} (L_j - L_i)^2. \tag{88}$$

This concludes the proof. □

## B.6 Proof of Theorem 6

*Proof.* Inspired by the median trick, we use the rescaling $\tilde{h} \ln N$ and define the $N \times N$ kernel matrix $K$ of which the $(i,j)$ element is defined as

$$K_{ij} := \exp\left(-\tilde{h} \ln N (4h_w)^{-2} (\ln p(x|\theta_i) - \ln p(x|\theta_j))^2\right). \tag{89}$$

Here, we introduced the additional bandwidth $\tilde{h} \ln N$ to rescale the kernel. For simplicity, define $-R_1 := \frac{1}{N^2 \tilde{h} \ln N} \sum_{i,j=1}^{N} \ln K_{ij}$.

By applying the Jensen inequality, we have

$$-R_1 = \frac{1}{N^2 \tilde{h} \ln N} \sum_{i,j=1}^{N} \ln K_{ij} \leq \frac{1}{\tilde{h} \ln N} \ln \sum_{i,j=1}^{N} \frac{K_{ij}}{N^2}. \tag{90}$$

Then note that from the definition of $K$, we have

$$\sum_{i,j=1}^{N} \frac{K_{ij}}{N^2} \leq 1. \tag{91}$$

We also have

$$-\tilde{h} R_1 \leq \frac{1}{\ln N} \ln \sum_{i,j=1}^{N} \frac{K_{ij}}{N^2} \leq \frac{1}{N} \ln \sum_{i,j=1}^{N} \frac{K_{ij}}{N^2} = -\frac{1}{N} \ln N^2 + \frac{1}{N} \ln \sum_{i,j=1}^{N} K_{ij}. \tag{92}$$

Then we define the new kernel function as

$$\tilde{K}_{ij} = K_{ij}^{1/2}. \tag{93}$$

We use the relation of the Frobenius norm and the trace,

$$\sum_{i,j=1}^{N} K_{ij} = \sum_{i,j} \tilde{K}_{i,j}^2 = \mathrm{Tr}(\tilde{K}^\top \tilde{K}). \tag{94}$$

By using a positive constant $\epsilon$, we obtain

$$\begin{aligned}
-\tilde{h} R_1 &\leq -\frac{1}{N} \ln N^2 + \frac{1}{N} \ln \mathrm{Tr}[\tilde{K} \tilde{K}^\top] \\
&\leq -\frac{1}{N} \ln N^2 + \frac{1}{N} \ln(\mathrm{Tr}[\tilde{K} \tilde{K}^\top] + \epsilon) \\
&\leq -\frac{1}{N} \ln N^2 - \frac{N-1}{N} \ln \epsilon + \frac{1}{N} \ln(\det[\epsilon I + \tilde{K} \tilde{K}^\top]) \\
&\leq -\frac{2}{N} \ln N - \frac{N-1}{N} \ln \epsilon + \frac{2}{N} \ln(\det[\epsilon^{1/2} I + \tilde{K}]).
\end{aligned} \tag{95}$$

From the second line to the third line, we used the following relation. We express the eigenvalues of $\tilde{K} \tilde{K}^\top$ as $\rho_i$s. Then since $\rho_i \geq 0$, we have

$$\begin{aligned}
&\det(\epsilon I + \tilde{K} \tilde{K}^\top) \\
&= \prod_i (\epsilon + \rho_i) \\
&\geq \epsilon^N + \prod_i \rho_i + \epsilon^{N-1} \sum_i \rho_i \\
&\geq \epsilon^N + \det[\tilde{K} \tilde{K}^\top] + \epsilon^{N-1} \mathrm{Tr}[\tilde{K} \tilde{K}^\top] \geq \epsilon^{N-1} \mathrm{Tr}[\tilde{K} \tilde{K}^\top],
\end{aligned} \tag{96}$$

and thus we have

$$\ln \mathrm{Tr}[\tilde{K} \tilde{K}^\top] \leq \ln \det(\epsilon I + \tilde{K} \tilde{K}^\top) - (N-1) \ln \epsilon. \tag{97}$$

Then apply this to the second line in Eq.(95). In the last inequality in Eq.(95), we used the relation

$$\begin{aligned}
\ln(\det[\epsilon^{1/2} I + \tilde{K}]) &= \frac{1}{2} \ln(\det[(\epsilon^{1/2} I + \tilde{K})(\epsilon^{1/2} I + \tilde{K})^\top]) \\
&= \frac{1}{2} \ln(\det[\epsilon I + \epsilon^{1/2}(\tilde{K} + \tilde{K}^\top) + \tilde{K} \tilde{K}^\top]) \\
&\geq \frac{1}{2} \ln(\det[\epsilon I + \tilde{K} \tilde{K}^\top]).
\end{aligned} \tag{98}$$

Thus, we have

$$-\tilde{h}R_1 \leq -\frac{2}{N}\ln N - \frac{N-1}{N}\ln \epsilon + \frac{2}{N}\ln(\det[\epsilon^{1/2}I + \tilde{K}])$$

$$\leq -\frac{2}{N}\ln N - \frac{N-1}{N}\ln \epsilon + \frac{2}{N}\sum_i^N \ln(\epsilon^{1/2} + \lambda_i)$$

$$\leq -\frac{N-1}{N}\ln \epsilon + \frac{2}{N}\sum_i^N \ln \frac{(\epsilon^{1/2} + \lambda_i)}{N}, \tag{99}$$

where $\lambda_i$ is the $i$-th eigenvalue of $\tilde{K}$.

By using the Gershgorin circle theorem [8] and the definition of $\tilde{K}$ for each $i$,

$$\lambda_i \leq \sum_j K_{ij}. \tag{100}$$

Thus, we need to upper-bound $\sum_j K_{ij}$ to estimate $\lambda_i$. For simplicity, we rescale $\epsilon^{1/2} \to \epsilon$. Recall the definition of the Gram matrix $K$. Given a positive constant $\epsilon$, $\tilde{h}$ is chosen such that $K$ satisfies following property:

$$\sum_j K_{ij} < N - \epsilon. \tag{101}$$

We discuss when $\epsilon$ that satisfies $-\ln \epsilon \leq 0$ and corresponding $\tilde{h}$ that satisfies $\sum_j K_{ij} < N - \epsilon$ exist later. If those $\epsilon$ and $\tilde{h}$ exist, we have

$$\lambda_i \leq N - \epsilon. \tag{102}$$

Then we have

$$\frac{(\epsilon^{1/2} + \lambda_i)}{N} \leq 1, \tag{103}$$

and thus we have

$$\frac{2}{N}\sum_i^N \ln \frac{(\epsilon^{1/2} + \lambda_i)}{N} \leq 0. \tag{104}$$

Thus, we have

$$-\tilde{h}R_1 \leq -\frac{N-1}{N}\ln \epsilon + \frac{2}{N}\sum_i^N \ln \frac{(\epsilon^{1/2} + \lambda_i)}{N}. \tag{105}$$

We would like to show that the right-hand side of the above is smaller than 0. Note that the second term is below 0. About the first term, if $\ln \epsilon \leq 0$, then we obtain

$$-\tilde{h}R_1 \leq -\frac{N-1}{N}\ln \epsilon + \frac{2}{N}\sum_i^N \ln \frac{(\epsilon^{1/2} + \lambda_i)}{N} \leq 0. \tag{106}$$

Next, we discuss when $\epsilon$ that satisfies $-\ln \epsilon \leq 0$ and corresponding $\tilde{h}$ that satisfies $\sum_j K_{ij} < N - \epsilon$ exist. If all the $N$ particles are exactly same, such $\epsilon$ and $\tilde{h}$ do not exist. However, since $R_1 = 0$ under that setting, we do not need to consider the existence of such $\epsilon$ and $\tilde{h}$.

Next, for $N$ particles, that is, $\theta_1, \ldots, \theta_N$, we assume that only $N-2$ particles are exactly the same. This means, for example $\theta_1 = \cdots = \theta_{N-2} \neq \theta_{N-1}, \theta_N$ (This includes the case when $\theta_{N-1} = \theta_N$). Then, for any $i$, we obtain $\sum_{j=1} K_{ij} < N-1$ if we choose $\tilde{h}$ sufficiently large. Thus, such $\tilde{h}$ exists. Also, this setting means that $\ln \epsilon = 0$, thus, $\epsilon$ that satisfies $-\ln \epsilon \leq 0$ and corresponding $\tilde{h}$ that satisfies $\sum_j K_{ij} < N - \epsilon$ exist. If $M \in (0, N-2]$ particles are the exactly the same, by setting $\tilde{h}$ sufficiently large, Eq.(106) holds in the same way as we comfirmed in the case of $M = N - 2$.

If $N-1$ particles are exactly the same, that is, for example $\theta_1 = \cdots = \theta_{N-1} \neq \theta_N$. We express $K_{1,N} = \epsilon'$. Then for any $i$, we have $\sum_j K_{ij} = N - (1 - \epsilon')$. This setting corresponds to $\epsilon = 1 - \epsilon'$. By setting $\tilde{h}$ sufficiently large, we can make $\epsilon'$ arbitrary small. This means $\epsilon \to 1^-$. Then, $\frac{N-1}{N} \ln \epsilon \to 0$ and $\frac{2}{N} \sum_i^N \ln \frac{(\epsilon^{1/2} + \lambda_i)}{N} \to \frac{2}{N} \sum_i^N \ln \frac{(1+\lambda_i)}{N} \leq 0$. Thus, Eq.(106) holds. In conclusion, given a constant $\epsilon$, if we use sufficiently large $\tilde{h}$, Eq.(106) holds.

In conclusion, if all the particles are not exactly the same, there exists $\epsilon$ such that Eq.(106) holds.

Finally, we get

$$\mathbb{E} \ln p(x|\theta) \leq \ln \mathbb{E} p(x|\theta) + \frac{2}{\tilde{h}N} \ln \det(\epsilon I + \tilde{K}) - \frac{2}{\tilde{h}N} \ln N \leq \ln \mathbb{E} p(x|\theta). \tag{107}$$

$\square$

### B.7 Proof of Eq.(23)

Our gram matrix $K$ is symmetric stationary kernel function, and thus it satisfies

$$\partial_{\theta_i} K_{i,j} = -\partial_{\theta_j} K_{i,j}. \tag{108}$$

Then from the log determinant property, we have

$$\partial_{\theta_i} \ln \det(\epsilon I + K) = \text{Tr}\left[(\epsilon I + K)^{-1} \partial_\theta K\right]. \tag{109}$$

Then we have

$$\begin{aligned}
&\partial_{\theta_i} \ln \det(\epsilon I + K) \\
&= \text{Tr}\left[(\epsilon I + K)^{-1} \partial_\theta K\right] \\
&= \sum_{i=1}^N (\epsilon I + K)^{-1}_{ij} \partial_{\theta_i} K_{ij} \\
&= -\sum_{i=1}^N (\epsilon I + K)^{-1}_{ij} \partial_{\theta_j} K_{ij},
\end{aligned} \tag{110}$$

and in the second equality, we used the definition of the trace. Then we get

$$\partial_{\theta_i} Obj(\{\theta_i\}) = \frac{1}{N} \partial_{\theta_i} \log p(x|\theta_i) + \frac{2}{\tilde{h}N} \sum_j (K + \epsilon I)^{-1}_{ij} \nabla_{\theta_i} K_{ij}. \tag{111}$$

Note that the small positive constant $c$ of GFSF shown in Table 1 is introduced so that $K + cI$ can have a inverse matrix.

### B.8 Proof of the repulsion to DPP

We define the new kernel function as

$$\tilde{G}_{ij} = G_{ij}^{1/2}, \tag{112}$$

where $G$ is defined in Eq.(17).

We use the relation of the Frobenius norm and the trace,

$$\sum_{i,j=1}^{N} G_{ij} = \sum_{i,j} \tilde{G}_{i,j}^2 = \text{Tr}(\tilde{G}^\top \tilde{G}). \tag{113}$$

From the definition of $G$ and $\tilde{G}$, we have $G_{ij} \leq 1$ and $\tilde{G}_{ij} \leq 1$. Thus, we have

$$\text{Tr}(\tilde{G}^\top \tilde{G}) \leq N^2. \tag{114}$$

This means that $\text{Tr}(\frac{\tilde{G}}{N}^\top \frac{\tilde{G}}{N}) \leq 1$. Since $G$ is the positive definite matrix, and from the above trace inequality, all the eigenvalues of $\frac{\tilde{G}}{N}$ is smaller than 1. Thus, from the trace inequality, we have

$$1 \geq \text{Tr}(\tilde{G}^\top \tilde{G})/N^2 \geq \det(\tilde{G}^\top \tilde{G})^{1/N}/N, \tag{115}$$

by taking the log, we have

$$0 \geq \ln \text{Tr}(\tilde{G}^\top \tilde{G})/N^2 \geq \frac{1}{N} \ln \det(\tilde{G}^\top \tilde{G}) - \ln N. \tag{116}$$

Finally, from Eq.(180)

$$\ln \mathbb{E}_{\rho_E(\theta)} p(x|\theta)$$
$$\geq \mathbb{E}_{\rho_E(\theta)} \ln p(x|\theta) - \frac{1}{N} \sum_{i=1}^{N} \ln \sum_{j=1}^{N} \frac{G_{ij}}{N}$$
$$\geq \mathbb{E}_{\rho_E(\theta)} \ln p(x|\theta) - \ln \sum_{i,j=1}^{N} \frac{G_{ij}}{N^2}$$
$$\geq \mathbb{E}_{\rho_E(\theta)} \ln p(x|\theta) - \ln \frac{\text{Tr}(\tilde{G}^\top \tilde{G})}{N^2}. \tag{117}$$

Then we use the following trace inequality, for the positive definite matrix $A$

$$\text{Tr}(I - A^{-1}) \leq \det A. \tag{118}$$

This inequality come from the fact that for a positive value $\rho$, we have

$$1 - \frac{1}{\rho} \leq \ln \rho. \tag{119}$$

Since the eigenvalues of $\frac{\tilde{G}^\top \tilde{G}}{N^2}$ is smaller than 1, $(I - \frac{\tilde{G}^\top \tilde{G}}{N^2})$ is the positive definite matrix. Thus the above inequality, we have

$$-\text{Tr}(\frac{\tilde{G}^\top \tilde{G}}{N^2}) \geq \ln \det(I - \frac{\tilde{G}^\top \tilde{G}}{N^2}) \geq \ln \det(I - \frac{\tilde{G}}{N}) + \ln \det(I + \frac{\tilde{G}}{N}) \geq \ln \det(I - \frac{\tilde{G}}{N}). \tag{120}$$

This concludes the proof.

# C  Additional PAC-Bayesian generalization error bounds

Here, we present the PAC-Bayesian bounds, which are related to w-SGLD, GFSF, and DPP. We can derive those bounds from the results in Appendix A.3 and the second-order Jensen inequalities.

## C.1 Ensemble PAC-Bayesian bound

Using the prior distribution introduced in Appendix A.3, we get the generalization error bound for the ensemble setting. For simplicity, we define

$$R_c(x, h_w) = \frac{1}{2(2h_w)^2} \frac{1}{N^2} \sum_{i,j=1}^{N} (\ln p(x|\theta_i) - \ln p(x|\theta_j))^2, \tag{121}$$

where

$$h_w^{-2} = \exp\left(2 \min_i \ln p(x|\theta_i) - 2 \max_j \ln p(x|\theta_j)\right). \tag{122}$$

**Theorem 10.** *For all $x, \theta$, $p(x|\theta) < \infty$ and for any prior distribution $\pi_{\mathrm{E}}$ over $\Theta_{\mathrm{E}}$ independent of $\mathcal{D}$ and for any $\xi \in (0,1)$ and $c > 0$, with probability at least $1 - \xi$ over the choice of training data $\mathcal{D} \sim \nu^{\otimes D}(x)$, for all probability distributions $\rho_E$ with $\mathrm{supp}(\rho_E) \subset \Theta_{\mathrm{E}}$, we have*

$$\mathrm{CE} \leq -\mathbb{E}_\nu[\mathbb{E}_{\rho_E} \ln p(x|\theta) + R_c(x, h_w)]$$

$$\leq -\frac{1}{D} \sum_{d=1}^{D} [\mathbb{E}_{\rho_E} \ln p(x_d|\theta) + R_c(x_d, h_w)] + \frac{\mathrm{KL}(\rho_E, \pi_{\mathrm{E}}) + \frac{\ln \xi^{-1} + \Psi'''_{\pi,\nu}(c,D)}{2}}{cD}, \tag{123}$$

where

$$\Psi'''_{\pi,\nu}(c, D) := \ln \mathbb{E}_{\pi(\theta,\theta')} \mathbb{E}_{\mathcal{D} \sim \nu^{\otimes D}(x)} e^{cD(-\mathbb{E}_{\nu(x)} L(x,\theta,\theta') + D^{-1} \sum_{d=1}^{D} L(x_d, \theta, \theta'))}, \tag{124}$$

and

$$L(x, \theta, \theta') := \ln p(x|\theta) + 2^{-1}(2h_w)^{-2}((\ln p(x|\theta))^2 - \ln p(x|\theta) \ln p(x|\theta')). \tag{125}$$

## C.2 Relation to w-SGLD

We define

$$R_w(x, G) := -\frac{1}{N} \sum_{i=1}^{N} \ln \sum_{j=1}^{N} \frac{G_{ij}}{N}, \tag{126}$$

where

$$G_{ij} := \exp\left(-8^{-1} h_w^{-2} (\ln p(x|\theta_i) - \ln p(x|\theta_j))^2\right). \tag{127}$$

Then, from the second-order Jensen inequality, we have

$$\ln \mathbb{E}_{\rho_{\mathrm{E}}(\theta)} p(x|\theta)$$
$$\geq \mathbb{E}_{\rho_{\mathrm{E}}(\theta)} \ln p(x|\theta) + R_c(x, h_w)$$
$$\geq \mathbb{E}_{\rho_{\mathrm{E}}(\theta)} \ln p(x|\theta) + R_w(x, G)$$
$$\geq \mathbb{E}_{\rho_{\mathrm{E}}(\theta)} \ln p(x|\theta). \tag{128}$$

Finally, we apply this to the result in Appendix C.1. Then we upper-bound Eq.(123) with the above inequality, and we obtain

**Theorem 11.** *For all $x, \theta$, $p(x|\theta) < \infty$ and for any prior distribution $\pi_{\mathrm{E}}$ over $\Theta_{\mathrm{E}}$ independent of $\mathcal{D}$ and for any $\xi \in (0,1)$ and $c > 0$, with probability at least $1 - \xi$ over the choice of training data $\mathcal{D} \sim \nu^{\otimes D}(x)$, for all probability distributions $\rho_E$ with $\mathrm{supp}(\rho_E) \subset \Theta_{\mathrm{E}}$, we have*

$$\mathrm{CE} \leq -\mathbb{E}_\nu[\mathbb{E}_{\rho_E} \ln p(x|\theta) + R_c(x, h_w)]$$

$$\leq -\frac{1}{D} \sum_{d=1}^{D} [\mathbb{E}_{\rho_E} \ln p(x_d|\theta) + R_w(x_d, G)] + \frac{\mathrm{KL}(\rho_E, \pi_{\mathrm{E}}) + \frac{\ln \xi^{-1} + \Psi'''_{\pi,\nu}(c,D)}{2}}{cD}, \tag{129}$$

where $\Psi'''$ is the same as Eq.(124).

## C.3 Relation to DPP

We define

$$R_D(x, \tilde{G}) := \frac{2}{N} \ln \det \tilde{G} - \ln N, \quad (130)$$

where

$$\tilde{G}_{ij} := \exp\left(-(4h_w)^{-2} (\ln p(x|\theta_i) - \ln p(x|\theta_j))^2\right). \quad (131)$$

Then, from the second-order Jensen inequality, we have

$$\begin{aligned}
&\ln \mathbb{E}_{\rho_E(\theta)} p(x|\theta) \\
&\geq \mathbb{E}_{\rho_E(\theta)} \ln p(x|\theta) + R_c(x, h_w) \\
&\geq \mathbb{E}_{\rho_E(\theta)} \ln p(x|\theta) + R_D(x, \tilde{G}).
\end{aligned} \quad (132)$$

Finally, we apply this to the result in Appendix C.1. Then we upper-bound Eq.(123) with the above inequality, and we obtain

**Theorem 12.** *For all $x, \theta$, $p(x|\theta) < \infty$ and for any prior distribution $\pi_E$ over $\Theta_E$ independent of $\mathcal{D}$ and for any $\xi \in (0,1)$ and $c > 0$, with probability at least $1 - \xi$ over the choice of training data $\mathcal{D} \sim \nu^{\otimes D}(x)$, for all probability distributions $\rho_E$ with $\mathrm{supp}(\rho_E) \subset \Theta_E$, we have*

$$\begin{aligned}
\mathrm{CE} &\leq -\mathbb{E}_\nu[\mathbb{E}_{\rho_E} \ln p(x|\theta) + R_c(x, h_w)] \\
&\leq -\frac{1}{D} \sum_{d=1}^D \left[ \mathbb{E}_{\rho_E} \ln p(x_d|\theta) + R_D(x_d, \tilde{G}) \right] + \frac{\mathrm{KL}(\rho_E, \pi_E) + \frac{\ln \xi^{-1} + \Psi'''_{\pi,\nu}(c,D)}{2}}{cD},
\end{aligned} \quad (133)$$

*where $\Psi'''$ is the same as Eq.(124).*

## C.4 Relation to GFSF

We define

$$R_g(x, K) := -\frac{2}{\tilde{h} N} \ln \det(\epsilon I + K) + \frac{2}{\tilde{h} N}, \quad (134)$$

where

$$K_{ij} := \exp\left(-\tilde{h} \ln N (4h_w)^{-2} (\ln p(x|\theta_i) - \ln p(x|\theta_j))^2\right), \quad (135)$$

and $\tilde{h}$ is a constant that is selected to satisfy the relation $\sum_{ij} K_{ij} < N^2 - \epsilon$.

Then, from the second-order Jensen inequality, we have

$$\begin{aligned}
&\ln \mathbb{E}_{\rho_E(\theta)} p(x|\theta) \\
&\geq \mathbb{E}_{\rho_E(\theta)} \ln p(x|\theta) + R_c(x, h_w) \\
&\geq \mathbb{E}_{\rho_E(\theta)} \ln p(x|\theta) + R_g(x, K) \\
&\geq \mathbb{E}_{\rho_E(\theta)} \ln p(x|\theta).
\end{aligned} \quad (136)$$

Finally, we apply this to the result in Appendix C.1. Then we upper-bound Eq.(123) with the above inequality, and we obtain

**Theorem 13.** *For all $x, \theta$, $p(x|\theta) < \infty$ and for any prior distribution $\pi_E$ over $\Theta_E$ independent of $\mathcal{D}$ and for any $\xi \in (0,1)$ and $c > 0$, with probability at least $1 - \xi$ over the choice of training data $\mathcal{D} \sim \nu^{\otimes D}(x)$, for all probability distributions $\rho_E$ with $\mathrm{supp}(\rho_E) \subset \Theta_E$, we have*

$$\begin{aligned}
\mathrm{CE} &\leq -\mathbb{E}_\nu[\mathbb{E}_{\rho_E} \ln p(x|\theta) + R_c(x, h_w)] \\
&\leq -\frac{1}{D} \sum_{d=1}^D [\mathbb{E}_{\rho_E} \ln p(x_d|\theta) + R_g(x_d, K)] + \frac{\mathrm{KL}(\rho_E, \pi_E) + \frac{\ln \xi^{-1} + \Psi'''_{\pi,\nu}(c,D)}{2}}{cD},
\end{aligned} \quad (137)$$

*where $\Psi'''$ is the same as Eq.(124).*

# D Comparison of our second-order Jensen inequality in Theorem 3 and that of the previous work [16, 12]

First, we discuss the difference of our loss function based second-order Jensen inequality and those of the previous work [16, 12] in terms of the derivation. Although both Our approach and previous work use the Taylor expansion up to a second order, the usage of the mean $\mu$ is different. First, let us consider the Taylor expansion of log up to a second-order around a constant $\mu$, then there exists a constant $g$ between $y$ and $\mu$ s.t.,

$$\ln y = \ln \mu + \frac{1}{\mu}(y-\mu) - \frac{1}{2g^2}(y-\mu)^2. \tag{138}$$

In the previous work [16, 12], given a random variable $Z$, they define $\mu := \mathbb{E}Z$, and $y := Z$. Then take the expectation. Then we have

$$\mathbb{E}\ln Z = \ln \mathbb{E}Z - \int_{\mathbb{R}^+} \frac{1}{2g(z)^2}(z-\mu)^2 p_Z(z)dz. \tag{139}$$

Note that the constant term and the second-order reminder term remain in the equation. Then by setting $Z := p(x|\theta)$, we get the second-order Jensen inequality of the previous work [16, 12]. We can see that the variance of the predictive distribution naturally appears since we define $\mu := \mathbb{E}p(x|\theta)$.

On the other hand, as we had seen in Appendix B.2, to derive our second-order inequality, we define $\mu = e^{\mathbb{E}\ln Z}$. Then from the Taylor expansion, we obtain

$$0 = \frac{1}{\mu}\mathbb{E}[Z - e^{\mathbb{E}\ln Z}] - \int_{\mathbb{R}^+} \frac{1}{2g(z)^2}(z-\mu)^2 p_Z(z)dz. \tag{140}$$

Compared to Eq.(139), in our Taylor expansion Eq.(140), the first-order term and reminder term remain in the equation. This results in the difference between our second-order Jensen inequalities and those of previous works.

Next, we discuss the difference of the second-order Jensen inequalities in terms of the meaning of the repulsions. In previous work [16], the second-order Jensen inequality was proved

$$\mathbb{E}_q \ln p(x|\theta) \leq \ln \mathbb{E}_q p(x|\theta) - V(x), \tag{141}$$

where

$$V(x) := (2\max_\theta p(x|\theta)^2)^{-1}\mathbb{E}_q\left[(p(x|\theta) - \mathbb{E}_q p(x|\theta))^2\right]. \tag{142}$$

Then by using Lemma 2, assuming that $x := e^{\ln p(x|\theta_i)}$ and $y := e^{\ln \mathbb{E}_q p(x|\theta)}$ and we get

$$(\ln p(x|\theta_i) - \ln \mathbb{E}_q p(x|\theta))^2 p(x|\theta_i)\mathbb{E}p(x|\theta) \leq (e^{\ln p(x|\theta_i)} - e^{\ln \mathbb{E}_q p(x|\theta)})^2. \tag{143}$$

Then we get

$$\mathbb{E}_q \ln p(x|\theta)$$
$$\leq \ln \mathbb{E}_q p(x|\theta) - V(x)$$
$$\leq \ln \mathbb{E}_q p(x|\theta) - \frac{1}{2\max_\theta p(x|\theta)^2}\mathbb{E}_q(\ln p(x|\theta_i) - \ln \mathbb{E}_q p(x|\theta))^2 p(x|\theta_i)\mathbb{E}_q p(x|\theta). \tag{144}$$

This expression is very similar to our Theorem 3, but it is different in a sense that this is not the weighted variance since $\ln \mathbb{E}_q p(x|\theta)$ appears. We cannot transform this to the mean of the loss function which is contrary to the Jensen inequality.

Thus our bound second order Jensen inequality in Theorem 3 and that of the previous work [16] is different bound, that is, ours focuses on $\mathbb{E} \ln p(x|\theta)$ and the previous work focuses on $\mathbb{E} p(x|\theta)$. We found that it is hard to claim that which is tighter. We believe it is interesting direction to study in what problems which bound is appropriate.

We numerically found that for the regression tasks and bandit problems, our approach consistently outperform the previous work. On the other hand, for classification tasks, it seems that both methods show almost equivalent performances. See Section 5.

# E Discussion about the repulsion force

## E.1 Transformation by the mean value theorem

As shown in the main paper, our loss repulsion can be translated to the parameter or model repulsion by using the mean value theorem. For example, there exist a parameter $\tilde{\theta}$ between $\theta_i$ and $\theta_j$ that is defined by a constant $t \in [0,1]$ s.t. $\tilde{\theta} := t\theta_i + (1-t)\theta_j$, which satisfied

$$\ln p(x; \theta_i) - \ln p(x; \theta_j) = \frac{\partial_{\tilde{\theta}} p(x; \tilde{\theta})}{p(x; \tilde{\theta})} \cdot (\theta_i - \theta_j), \tag{145}$$

and similar relation also holds for $\ln p(y|f(x; \theta_i)) - \ln p(y|f(x; \theta_j))$. Thus, we can transform our loss repulsion to parameter or model repulsion. From

$$\| \ln p(x; \theta_i) - \ln p(x; \theta_j) \|^2 = \| \frac{\partial_{\tilde{\theta}} p(x; \tilde{\theta})}{p(x; \tilde{\theta})} \cdot (\theta_i - \theta_j) \|^2, \tag{146}$$

and neglecting the bandwidth for simplicity, we define a Gram matrix

$$K_{ij} := \exp\left(-\| \ln p(x; \theta_i) - \ln p(x; \theta_j) \|^2\right) = \exp\left(-\| \frac{\partial_{\tilde{\theta}} p(x; \tilde{\theta})}{p(x; \tilde{\theta})} \cdot (\theta_i - \theta_j) \|^2\right). \tag{147}$$

Then by taking the partial derivative with respect to $\theta_i$,

$$\partial_{\theta_i^{(d)}} K_{ij} = -2(\theta_i^{(d)} - \theta_j^{(d)}) \frac{\partial_{\theta_{\tilde{(d)}}} p(x; \tilde{\theta})}{p(x; \tilde{\theta})} K_{ij} + \text{Corr}, \tag{148}$$

where $\theta_i^{(d)}$ corresponds to the $d$-th dimension of the parameter $\theta_i$. The first term in the above corresponds to the parameter repulsion and the second term is the correction term. We can get the similar relation to the model repulsion. However, it is difficult to obtain the explicit form of $\frac{\partial_{\tilde{\theta}} p(x; \tilde{\theta})}{p(x; \tilde{\theta})}$.

## E.2 Model repulsion

**Regression**

We first discuss the relation to model repulsion. For a regression problem, we assume that $p(y|f(x; \theta))$ is the Gaussian distribution with unit variance for simplicity,

$$\ln p(y|f(x;\theta)) = -\frac{1}{2}(y - f(x;\theta))^2 + const. \tag{149}$$

For f-PVIs, assume that $K_{ij} = \exp(-\frac{1}{2h^2} \| f_i - f_j \|^2)$ where the bandwidth is $h$. Then the model repulsion is expressed as

$$\partial_{\theta_i} K(f_i, f_j) = -\frac{1}{h^2}(f_i - f_j) K_{i,j} \partial_{\theta_i} f_i. \tag{150}$$

On the other hand, the kernel function of our loss repulsion is from Eq.(17)

$$G_{ij} := \exp\left(-(8h_w^2)^{-1}\|\ln p(y|f(x;\theta_i)) - \ln p(y|f(x;\theta_j)\|^2\right)$$
$$= \exp\left(-(8h_w^2)^{-1}\frac{1}{4}\|f(x;\theta_i) - f(x;\theta_j)\|^2\|f(x;\theta_i) + f(x;\theta_j) - 2y\|^2\right). \quad (151)$$

We define $L(f_i) := \ln p(y|f(x;\theta_i))$ and $dL_{ij} := \partial_{f_i}L(f_i) + \partial_{f_j}L(f_j)$. The derivative of the Gram matrix $G$ is expressed as

$$\partial_{\theta_i} G_{ij} = -(\underbrace{(f_i - f_j)\|dL_{ij}\|^2}_{i)} + \underbrace{\partial_{f_i}L(f_i)dL_{ij}\|f_i - f_j\|^2}_{ii)})(4h_w)^{-2}G_{ij}\partial_{\theta_i} f_i. \quad (152)$$

As we discussed in the main part, the first term corresponds to the model repulsion and the second term corresponds to the correction term. Thus our loss repulsion is closely related to model repulsion.

We can further simplify the above relation as follows. We define $l(f_i, f_j) := \|f_i + f_j - 2y\|^2$. We define a constant $l_0$ as a constant that satisfies $l_0 \leq \min_{(i,j)} l(f_i, f_j)$, we get

$$G_{i,j} \leq \exp\left(-(8h_w^2)^{-1}\frac{l_0}{4}\|f_i - f_j\|^2\right) := G_{i,j}^0. \quad (153)$$

Then by taking the partial derivative

$$\partial_{\theta_i} G_{i,j}^0 = -(4h_w)^{-2} l_0 (f_i - f_j) G_{i,j}^0 \partial_{\theta_i} f_i, \quad (154)$$

and thus, this repulsion force corresponds to the f-PVIs.

**Classification**

For classification task, if the class number is C, then, in standard models

$$p(x|\theta) := \text{Multinomial}(y|\text{softmax}(f(x;\theta))), \quad (155)$$

and $f$ is a C-dimensional output neural network. We assume that there is $N$ number of ensembles. We express $f_i^c := f^c(x, \theta_i)$ as the $c$-th output of the $i$-th neural network. We express the output of the softmax function as $\{p_i^{c'}\}_{c'=1}^c$. Then if the true class label is $y = t$, we have

$$\ln p(y|f(x;\theta_i)) = f_i^t - \ln \sum_c f_i^c. \quad (156)$$

First of all, we consider the model repulsion of f-PVI. Assume that $K_{ij} = \exp(-\frac{1}{2h^2}\|f_i - f_j\|^2)$ and the model repulsion of f-SVGD is expressed as

$$-\frac{1}{h^2}\sum_{c'=1}^c (f_i^{c'} - f_j^{c'}) K_{ij} \partial_{\theta_i} f_i^{c'}. \quad (157)$$

Then, we directly calculate the derivative of the loss repulsion and connect it to the model repulsion.

We can write the $(i, j)$-th element of the gram matrix as

$$G_{i,j} := \exp\left(-(8h_w^2)^{-1}(\ln p(y|f(x;\theta_i)) - \ln p(y|f(x;\theta_j)))^2\right)$$
$$= \exp\left(-(8h_w^2)^{-1}\|(f_i^t - f_j^t - (\ln \sum_c f_i^c - \ln \sum_c f_j^c)\|^2\right). \quad (158)$$

Then by calculating the derivative of the Gram matrix, we have

$$\partial_{\theta_i} G_{ij} = -\sum_{c'=1}^{c} (\underbrace{(f_i^{c'} - f_j^{c'})}_{i)} - \underbrace{(Z_i^{c',t} - Z_j^{c',t})}_{ii)})(\delta_{c',t} - p_i^{c'})(2h_w)^{-2} G_{ij} \partial_{\theta_i} f_i^{c'}, \quad (159)$$

where

$$\delta_{c',t} = \begin{cases} 1 & (c' = t) \\ 0 & (c' \neq t), \end{cases} \quad (160)$$

and

$$Z_i^{c',t} = \ln \sum_{c''=1}^{c} e^{f_i^{c''} - f_i^{c'} \delta_{c'' \neq t}}, \quad (161)$$

where

$$\delta_{c'' \neq t} = \begin{cases} 1 & (c'' \neq t) \\ 0 & (c'' = t). \end{cases} \quad (162)$$

Thus, in Eq.(159), similary to the regression setting, the first term $i)$ corresponds to the model repulsion and the second term is the correction term.

For simplicity, we define $f_i := f(x; \theta_i)$. And we define $d_1(f_i^c, f_j^c) := |f_i^c - f_j^c|$ and $d_2(f_i, f_j) := |\ln \sum_c f_i^c - \ln \sum_c f_j^c|$.

$$G_{i,j} \leq \exp\left(-(8h_w^2)^{-1}\|f_i - f_j\|^2 + 2(8h_w^2)^{-1} d_1(f_i^c, f_j^c) d_2(f_i, f_j)\right) := \tilde{G}_{i,j}. \quad (163)$$

Moreover, we define a constant $d_1^0$ such that $d_1^0 \leq \max_{(c,i,j)} |f_i^c - f_j^c|$ and define a constant $d_2^0$ such that $d_2^0 \leq \max_{(i,j)} |\ln \sum_c f_i^c - \ln \sum_c f_j^c|$, we define

$$G_{i,j}^0 := \tilde{c} \exp\left(-(8h_w^2)^{-1}\|f_i - f_j\|^2\right), \quad (164)$$

where $\tilde{c} := \exp\left(2(8h_w^2)^{-1} d_1^0 d_2^0\right)$. These satisfies

$$G_{i,j} \leq \tilde{G}_{i,j} \leq G_{i,j}^0. \quad (165)$$

Then by taking the partial derivative

$$\partial_{\theta_i} \tilde{G}_{i,j} = -(2h_w)^{-2}(f_i^t - f_j^t) G_{i,j} \partial_{\theta_i} f_i^t. \quad (166)$$

Thus, our loss repulsion is closely related to model repulsion.

### E.3 Parameter repulsion

We found that it is difficult to derive the parameter repulsion without using the mean value theorem. On the other hand, when we are allowed to use the reference repulsion shown below, then we can derive the parameter repulsion To introduce the parameter repulsion force, we use the following

**Lemma 4.** *For any reference probability density $p(\theta)$ that is bounded below $0 \leq p(\theta) < M$, under the same assumptions with Collorary 1, we have*

$$\ln \mathbb{E} p(x|\theta) \geq \mathbb{E} \ln[p(x|\theta) p(\theta)] + R(x, \theta) + \frac{1}{8h^2 N^2} \sum_{i,j=1}^{N} (p(\theta_i) - p(\theta_j))^2 - \ln M. \quad (167)$$

*where $R$ is the same as Theorem 3.*

*Proof.* From our second order Jensen inequality, we have

$$M \geq \ln \mathbb{E}_{\rho_E(\theta)}[p(\theta)] \geq \mathbb{E}_{\rho_E(\theta)}[\ln p(\theta)] + \frac{1}{8h_\omega^2 N^2} \sum_{i,j=1}^{N} (p(\theta_i) - p(\theta_j))^2. \qquad (168)$$

Then by combining this with Eq.(16), Lemma 4 is proved. □

This lemma state that by introducing the reference distribution $p(\theta)$, we obtain the lower bound and repulsion term based on $p(\theta)$. As we did in the main paper, we can lower bound the variance of log of prior distribution in various ways. And assume that we lower bound it in gram matrix form

$$K_{i,j} := \exp\left(-(8h^2)^{-1}(\ln p(\theta_i) - \ln p(\theta_j))^2)\right). \qquad (169)$$

Assume that $p(\theta)$ is a exponential family distribution,

$$\ln p(\theta) = \eta u(\theta) + Const, \qquad (170)$$

where $\eta$ is a natural parameter and $u(\theta)$ is a sufficient statistics. Then we have

$$K_{i,j} := \exp\left(-(8h^2)^{-1}\eta^2(u(\theta_i) - u(\theta_j))^2)\right). \qquad (171)$$

Then by taking the partial derivative, we have

$$\partial_{\theta_i} K_{i,j} = -8h^{2^{-1}}\eta^2(u(\theta_i) - u(\theta_j))K_{i,j}\partial_{\theta_i} u(\theta_i). \qquad (172)$$

In standard PVIs, the parameter repulsion force is

$$\partial_{\theta_i} K_{i,j} = -(8h^2)^{-1}(\theta_i - \theta_j)K_{i,j}. \qquad (173)$$

In order to discuss these repulsions, we assume that $p(\theta)$ is a standard Gaussian distribution. Then by taking the partial derivative, we have

$$\partial_{\theta_i} K_{i,j} = -h^2(\theta_i - \theta_j)d_4(i,j)K_{i,j} - h^2(\theta_i - \theta_j)^2 \partial_{\theta_i} d_4(i,j)K_{i,j}. \qquad (174)$$

where $d_4(i,j) := |\theta_i + \theta_j|^2$.

On the other hand, and if there exists a constant $d_4^0$ such that $d_4^0 \leq \min_{i,j} d_4(i,j)$. Then

$$K_{ij} \leq K_{i,j}^0 := \exp\left(-h^2 d_4^0 (\theta_i - \theta_j)^2)\right). \qquad (175)$$

and by taking the partial derivative, we have

$$\partial_{\theta_i} K_{i,j}^0 = -h^2(\theta_i - \theta_j)d_4^0(i,j)K_{i,j}^0. \qquad (176)$$

Note that the prior repulsion force is the same as the repulsion force of PVIs of Eq.(173).

### E.4 Data summation inside the variance

In the existing model repulsion force, for example, regression tasks, the kernel function $K$ is defined as

$$K_{ij} := \exp(-\|f_i(\boldsymbol{x}) - f_j(\boldsymbol{x})\|^2) = \exp(-(\|f_i(x_1) - f_j(x_1)\|^2 + \ldots + \|f_i(x_b) - f_j(x_b)\|^2)), \qquad (177)$$

where $b$ is the minibatch size. Thus, we take the summation with respect to the data points.

Compared to this model repulsion, our loss repulsion is, for example, expressed as

$$-\frac{1}{D}\sum_{d=1}^{D}\frac{1}{2(2h_\omega)^2 N^2}\sum_{ij}\|\ln p(x_d|\theta_i) - \ln p(x_d|\theta_j)\|^2. \tag{178}$$

As defined in Section 3.2, when we define our Gram matrix $G$, we can incorporate the data summation term inside the Gram matrix as

$$G_{ij} := \exp\left(-(8h_w^2)^{-1}\sum_{d=1}^{D}(\ln p(x_d|\theta_i) - \ln p(x_d|\theta_j))^2\right). \tag{179}$$

Applying the Jensen inequality, we obtain

$$\frac{1}{D}\sum_{d=1}^{D}\ln \mathbb{E}_{\rho_E(\theta)}p(x_d|\theta) \geq \frac{1}{D}\sum_{d=1}^{D}\mathbb{E}_{\rho_E(\theta)}\ln p(x_d|\theta) - \frac{1}{DN}\sum_{i=1}^{N}\ln\sum_{j=1}^{N}\frac{G_{ij}}{N}$$
$$\geq \frac{1}{D}\sum_{d=1}^{D}\mathbb{E}_{\rho_E(\theta)}\ln p(x_d|\theta). \tag{180}$$

# F  Relation to the misspecified model setting

In the previous work [16], the advantage of the second-order Jensen inequality was analyzed in the case of the misspecified model setting, that is, for any $\theta$, $\nu(x) \neq p(x|\theta)$. They proved the following theorems:

**Theorem 14.** *[16] Let us denote $\theta_{\mathrm{ML}}^* := \arg\min_{\theta} \mathrm{KL}(\nu(x), p(x|\theta))$ and $p_{\mathrm{ML}}$ is the distribution obtained by minimizing $\mathbb{E}_{p(\theta),\nu(x)}[-\ln p(x|\theta)]$. Then $p_{\mathrm{ML}}$ also minimizes $\mathrm{CE}(p) := \mathbb{E}_{\nu(x)}[-\ln \mathbb{E}_{p(\theta)} p(x|\theta)]$ if and only if for any distribution $p$ over $\Theta$ we have that*

$$\mathrm{KL}(\nu(x), p(x|\theta_{\mathrm{ML}}^*)) \leq \mathrm{KL}(\nu(x), \mathbb{E}_p p(x|\theta)). \tag{181}$$

*and $p_{\mathrm{ML}}$ can always be characterized as a Dirac distribution center around $\theta_{\mathrm{ML}}^*$, tha is, $p_{\mathrm{ML}} = \delta_{\theta_{\mathrm{ML}}^*}(\theta)$.*

According to this theorem, the previous work [16] claimed that Bayesian posterior distribution is an optimal strategy under perfect model speccification since $\mathrm{KL}(\nu(x), p(x|\theta_{\mathrm{ML}}^*)) = 0$ and $p_{\mathrm{ML}}^*$ minimizes $\mathrm{CE}(p)$.

However, in many practical settings, we cannot expect the perfect model specification. Then under the misspecified model settings, the previous work [16] clarified that the second order Jensen inequality provides the better solution as follows:

**Theorem 15.** *[16] Let us denote the $p_V^*$ as the distribution obtained by minimizing $\mathbb{E}_{p,\nu(x)}[-\ln p(x|\theta)] - \mathbb{E}_{\nu(x)}V(x)$ and $p_{\mathrm{ML}}$ is the distribution obtained by minimizing $\mathbb{E}_{p,\nu(x)}[-\ln p(x|\theta)]$. Then following inequality holds,*

$$\mathrm{KL}(\nu(x), \mathbb{E}_{p_V^*} p(x|\theta)) \leq \mathrm{KL}(\nu(x), \mathbb{E}_{p_{\mathrm{ML}}} p(x|\theta)). \tag{182}$$

*Here the equality holds if we are under perfect model specification, that is, there exists a parameter $\theta^*$ that satisfies $\nu(x) = p(x|\theta^*)$.*

Thus, this theorem clarifies that under model misspecified setting, learning the second-order Jensen inequality can be a better strategy than Bayesian inference.

Motivated these previous results, we can show the similar inequality for our loss function based second-order Jensen inequality:

**Theorem 16.** *Let us denote the $p_R^*$ as the distribution obtained by minimizing $\mathbb{E}_{p,\nu(x)}[-\ln p(x|\theta)] - \mathbb{E}_{\nu(x)}R(x)$ and $p_{\text{ML}}$ is the distribution obtained by minimizing $\mathbb{E}_{p,\nu(x)}[-\ln p(x|\theta)]$. Then following inequality holds,*

$$\text{KL}(\nu(x), \mathbb{E}_{p_R^*}p(x|\theta)) \leq \text{KL}(\nu(x), \mathbb{E}_{p_{\text{ML}}}p(x|\theta)). \tag{183}$$

*and the equality holds if we are under perfect model specification.*

The proof of this theorem is exactly the same as that of Theorem 15[16]. Here, we show the outline of the proof.

*Proof.* Define $\Omega$ as the space of distributions $p$ over $\Theta$ that satisfies $\mathbb{E}_{\nu(x)}R(x) = 0$. Then we have

$$\min_{p \in \Omega} \mathbb{E}_{p,\nu(x)}[-\ln p(x|\theta)] - \mathbb{E}_{\nu(x)}R(x) = \min_{p \in \Omega} \mathbb{E}_{p,\nu(x)}[-\ln p(x|\theta)] = \mathbb{E}_{p_{\text{ML}},\nu(x)}[-\ln p(x|\theta)], \tag{184}$$

where the last equality comes from Lemma A.6 [16] (just replace V(x) with R(x)). Then we have

$$\mathbb{E}_{p_R^*,\nu(x)}[-\ln p(x|\theta)] - \mathbb{E}_{\nu(x)}R(x) \leq \mathbb{E}_{p_{\text{ML}},\nu(x)}[-\ln p(x|\theta)]. \tag{185}$$

Here the left-hand side is the minimum of the second order Jensen inequality for all the distributions $p$ over $\Theta$ and right-hand is the minimum within $\Omega$. Then from the second-order Jensen inequality, we have

$$\text{CE}(p_R^*) \leq \mathbb{E}_{p_{\text{ML}},\nu(x)}[-\ln p(x|\theta)]. \tag{186}$$

From lemma A.6 [16],

$$\mathbb{E}_{p_{\text{ML}},\nu(x)}[-\ln p(x|\theta)] = \text{CE}(p_{\text{ML}}), \tag{187}$$

thus we get

$$\text{CE}(p_R^*) \leq \text{CE}(p_{\text{ML}}). \tag{188}$$

By adding the entropy of $\nu(x)$, we get the inequality of the Theorem.

Next we study when the equality holds. From Theorem 14[16], under the perfect model specification, we have that for any $p$ over $\Theta$, we have

$$\text{CE}(p_{\text{ML}}) \leq \text{CE}(p). \tag{189}$$

Combining this with Eq.(188), we obtain

$$\text{CE}(p_R^*) = \text{CE}(p_{\text{ML}}), \tag{190}$$

under perfect model specification. □

# G  Numerical experiments

In this section, we describe the detail settings of the experiments. We also present the additional experimental results.

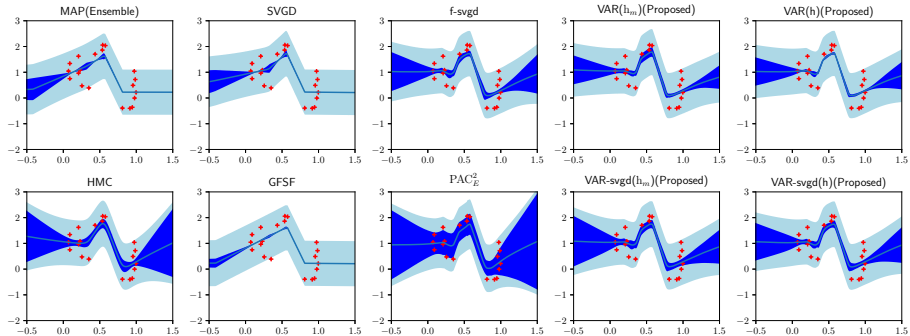

Figure 2: Uncertainty of the regressions. Blue line is the predictive mean, dark shaded area is the epistemic uncertainty from $\rho_E(\theta)$, and light shaded area is aleatory uncertainty that comes from $p(x|\theta)$.

### G.1 Smoothing the gradient

As for KL divergence, since $\rho_E$ is an empirical measure, we need to choose a certain prior distribution as introduced previously [16] (see Appendix A.3 for details). To eliminate such a limitation, we smooth the gradient of $L(\{\theta_i\})$ using SVGD. We call this approach as *Var-svgd*. See Appendix G for the explicit expression of the each method. In Var-SVGD, the update equation is given as

$$\theta_i^{\text{new}} \leftarrow \theta_i^{\text{old}} + \eta \frac{1}{N} \sum_{j=1}^{N} K_{ij} \partial_{\theta_j^{\text{old}}} \left( \log p(\mathcal{D}|\theta_j^{\text{old}}) \pi(\theta_j^{\text{old}}) + R(\mathcal{D}, h) \right) + \partial_{\theta_j^{\text{old}}} K_{ij}, \quad (191)$$

where $R(\mathcal{D}, h)$ is $\sum_{i=1}^{D} R(x_i, h)$, and $K_{ij}$ is the Gaussian kernel defined in the same way as the model repulsion, of which bandwidth is tuned by the median trick.

### G.2 Toy data experiments

#### G.2.1 Detail settings of the main paper

The setting is the same as that of the previous work of f-PVI [24]. We generated the data by $y = x + \sin 4(x + \epsilon) + \sin 13(x + \epsilon) + \epsilon$, $\epsilon \sim N(0, 0.0009)$. We generated $x$ as follows: 12 points are drawn from Uniform$(0, 0.6)$ and 8 points from Uniform$(0, 0.8)$. We used the Adam optimizer with a learning rate of 0.001. We fixed the observation variance $N(y|f(x;\theta))$ during the optimization with 0.2. In addition to the main paper, here, we also show the result of Var-svgd and GFSF.

#### G.2.2 Additional toy data experiments of the regression task

Here, we consider the linear regression problem for toy data experiment, especially focuses on the model misspecified setting discussed in Appendix F.

We generated the toy data following $y = x + 1 + \epsilon$ where $\epsilon \sim N(0, 5)$ and $x \sim$ Uniform$(-10, 10)$ and thus this is the one-dimensional regression task. As a model, we prepared $y = \theta_1 x + \theta_2 + \epsilon'$ where $\epsilon \sim N(0, 1)$. We used the standard Gaussian priors for each $\theta$s. Thus, this is a model misspecified setting discussed in Appendix F. We used 10 particles and optimize them in the framework of MAP, SVGD, PAC$_E^2$ and our proposed approach. We optimized each model by Adam with stepsize 0.001. We also show the result of the HMC, which is the baseline method in Bayesian inference. The result is shown in Figure 3, which visualizes 95% credible intervals for the prediction and mean estimate corresponding to aleatoric and epistemic uncertainties. Shown in the figure, SVGD shows almost similar uncertainty as HMC. Note that HMC visualize the uncertainty of Bayesian inference. On the other hand, the second order method of ours and $PAC_E^2$ show larger epistemic uncertainty than those of HMC and SVGD. Note that in the result of $PAC_E^2$, most training data points are inside the 95% credible intervals. In Table 5, we compared the quality of fittings of the models. As we can see, except for $PAC_E^2$, the fitting qualities are almost equivalent. Thus, as we discussed in the main paper, there is a trade-off between model fitting and enhancing diversity in the ensemble learning in the second order Jensen inequality. The results of HMC, f-SVGD seem small diversity since they are based on Bayesian inference, while $PAC_E^2$ shows sufficient diversity. On the other hand, the quality of model fitting of Bayesian inference seems superior to that of $PAC_E^2$. It seems that our proposed Var and Var-svgd seem the intermediate performance and diversity between Bayesian inference and $PAC_E^2$.

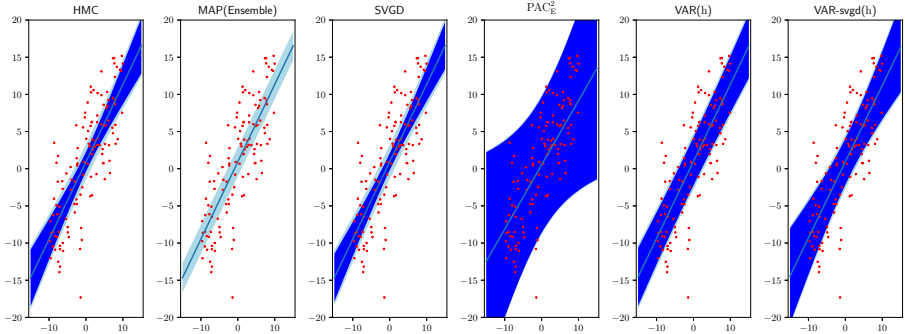

Figure 3: Uncertainty of the regressions. Blue line is the predictive mean, dark shaded area is the epistemic uncertainty from $\rho_E(\theta)$, and light shaded area is aleatory uncertainty that comes from $p(x|\theta)$.

Table 5: Comparison of performances

| Test Accuracy | | | | | | Test log likelihood | | | | | |
|---|---|---|---|---|---|---|---|---|---|---|---|
| HMC | MAP | f-SVGD | $PAC_E^2$ | Cov(h) | Cov-svgd(h) | HMC | MAP | f-SVGD | $PAC_E^2$ | Cov(h) | Cov-svgd(h) |
| 5.089 | 5.079 | 5.079 | 5.220 | 5.075 | 5.070 | 3.290 | 3.171 | 3.171 | 3.310 | 3.180 | 3.190 |

### G.3 Regression

The setting is the same as that of the previous work of f-PVI [24]. We used the Adam optimizer with learning rate 0.004. We used a batch size of 100 and run 500 epochs for the dataset size $D$ is smaller than 1000. For larger dataset, we used batch size of 1000 and run 3000 epochs. The result in the table is the 10 repetition except for Protein data which is the result of 5 repetition. In addition to the main paper, here, we also show the results of Var-svgd in Table 6,7.

### G.4 Classification

In addition to the main paper, we show the robustness to the adversarial samples in Figure 4. The experimental settings are exactly the same as that of the previous work [24].

We optimized the parameters using Adam with stepsize 0.0005. We used a batch-size with 1000 and run 1000 epochs. For MNIST adversarial experiments, we used a feed-forward network with ReLu activation and 3 hidden layers with 1000 units. The hyperparameter settings are the same as the result of the main paper. To generate attack samples, we used the iterative fast gradient sign method (I-FGSM) to generate the attack samples. We restrict the update with $l^\infty$ norm of the perturbation step 0.01.

Table 6: Benchmark results on test RMSE for the regression task

| Dataset | Avg. Test RMSE | | | | | | | |
|---|---|---|---|---|---|---|---|---|
| | MAP | SVGD | $PAC_E^2$ | f-SVGD | VAR(h) | VAR-svgd(h) | VAR($h_m$) | VAR-svgd($h_m$) |
| Concrete | 5.19±0.3 | 5.21±0.4 | 5.49±0.3 | 4.32±0.1 | 4.33±0.1 | 4.35±0.2 | 4.36±0.2 | 4.27±0.4 |
| Boston | 2.98±0.4 | 2.71±0.6 | 4.03±0.5 | 2.54±0.3 | 2.54±0.3 | 2.52±0.3 | 2.52±0.3 | 2.53±0.4 |
| Wine | 0.65±0.04 | 0.63±0.03 | 1.03±0.09 | 0.61±0.03 | 0.61±0.03 | 0.61±0.03 | 0.61±0.03 | 0.61±0.03 |
| Power | 3.94±0.03 | 3.90±0.14 | 5.04±0.21 | 3.77±0.03 | 3.76±0.03 | 3.40±0.05 | 3.76±0.06 | 3.75±0.08 |
| Yacht | 0.86±0.05 | 0.83±0.10 | 0.70±0.21 | 0.59±0.09 | 0.59±0.09 | 0.58±0.12 | 0.59±0.09 | 0.59±0.10 |
| Protein | 4.61±0.02 | 4.22±0.09 | 4.17±0.05 | 3.98±0.03 | 3.95±0.05 | 3.93±0.07 | 3.96±0.06 | 3.93±0.04 |

Table 7: Benchmark results on test negative log likelihood for the regression task

| Dataset | Avg. Test negative log likelihood | | | | | | | |
|---|---|---|---|---|---|---|---|---|
| | MAP | SVGD | f-SVGD | $PAC_E^2$ | VAR(h) | VAR-svgd(h) | VAR($h_m$) | VAR-svgd($h_m$) |
| Concrete | 3.11±0.12 | 3.11±0.14 | 3.16±0.10 | 2.86±0.02 | 2.82±0.09 | 2.80±0.06 | 2.87±0.09 | 2.81±0.06 |
| Boston | 2.62±0.2 | 2.49±0.4 | 2.61±0.3 | 2.46±0.1 | 2.39±0.2 | 2.35±0.2 | 2.48±0.4 | 2.41±0.2 |
| Wine | 0.97±0.07 | 0.96±0.06 | 1.26±0.02 | 0.90±0.05 | 0.89±0.04 | 0.89±0.06 | 0.89±0.04 | 0.90±0.07 |
| Power | 2.79±0.05 | 2.78±0.03 | 3.17±0.01 | 2.76±0.05 | 2.79±0.03 | 2.79±0.03 | 2.76±0.02 | 2.76±0.03 |
| Yacht | 1.23±0.05 | 1.32±0.6 | 0.80±0.4 | 0.96±0.3 | 0.87±0.3 | 0.81±0.2 | 1.03±0.3 | 0.91±0.2 |
| Protein | 2.95±0.00 | 2.86±0.02 | 2.84±0.01 | 2.80±0.01 | 2.81±0.01 | 2.80±0.01 | 2.80±0.01 | 2.80±0.01 |

Table 8: Benchmark results on test accuracy and negative log likelihood for the classification task

| Dataset | Test Accuracy | | | | | | | Test log likelihood | | | | | | |
|---|---|---|---|---|---|---|---|---|---|---|---|---|---|---|
| | MAP | $PAC_E^2$ | f-SVGD | VAR(h) | VAR-svgd(h) | VAR($h_m$) | VAR-svgd($h_m$) | MAP | $PAC_E^2$ | f-SVGD | VAR(h) | VAR-svgd(h) | VAR($h_m$) | VAR-svgd($h_m$) |
| Mnist | 0.981 | 0.986 | 0.987 | 0.988 | 0.988 | 0.988 | 0.988 | 0.057 | 0.042 | 0.043 | 0.040 | 0.041 | 0.041 | 0.041 |
| Cifar 10 | 0.935 | 0.919 | 0.927 | 0.929 | 0.928 | 0.927 | 0.924 | 0.215 | 0.270 | 0.241 | 0.238 | 0.240 | 0.242 | 0.242 |

Table 9: Cumulative regret relative to that of the Uniform sampling.

| Dataset | MAP | $PAC_E^2$ | f-SVGD | VAR(h) | VAR-svgd(h) | VAR($h_m$) | VAR-svgd($h_m$) |
|---|---|---|---|---|---|---|---|
| Mushroom | 0.129±0.098 | 0.037±0.012 | 0.043±0.009 | **0.029±0.010** | 0.037±0.012 | 0.036±0.012 | 0.038±0.010 |
| Financial | 0.791±0.299 | 0.189±0.025 | 0.154±0.017 | 0.155±0.024 | 0.176±0.023 | **0.128±0.017** | 0.153±0.020 |
| Statlog | 0.675 ±0.287 | 0.032±0.0025 | 0.010±0.0003 | **0.006±0.0003** | 0.007±0.0004 | 0.008±0.0005 | 0.011±0.004 |
| CoverType | 0.610±0.051 | 0.396±0.006 | 0.372±0.007 | **0.289±0.003** | 0.320±0.005 | 0.343±0.002 | 0.369±0.004 |

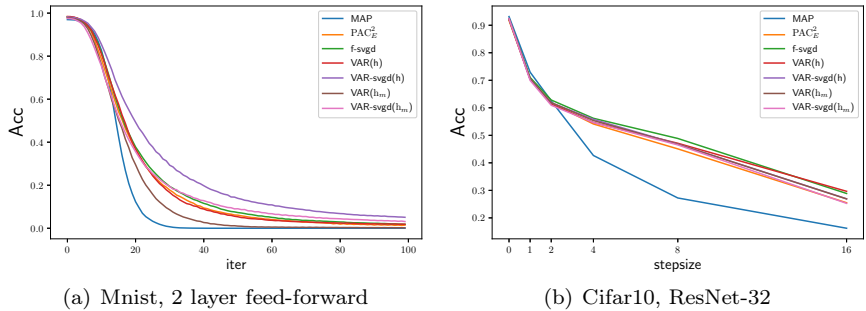

(a) Mnist, 2 layer feed-forward  (b) Cifar10, ResNet-32

Figure 4: Out of distribution performances

For Cifar10 experiments, we used ResNet32 and optimized Momentum sgd with the stepsize 0.09. We used a batchsize with 128 and run 200 epochs. We generated attack samples using FGSM under different stepsizes.

The result is shown in Figure 4. For both experiments, f-SVGD, $PAC_E^2$, and our proposed methods showed the more robustness against adversarial samples than that of the MAP estimate.

In addition to the main paper, here, we also show the results of Var-svgd in Table 8.

### G.5 Contextual bandit

First, we describe the problem setting. First we are given a context set $\mathcal{S}$. For each time step, $t = 1, \ldots, T$, a context $s_t \in \mathcal{S}$ is provided to a agent from the environment. Then, the agent choose the action $a_t$ from the available set $a_t \in \{1, \ldots, A\}$ based on the context $s_t$ and get a reward $r_{a_t, t}$. The goal of contextual bandit problem is to minimize the pseudo regret

$$R_T = \max_{g:\mathcal{S} \to \{1,\ldots,A\}} \mathbb{E}\left[\sum_{t=1}^{T} r_{g(s_t),t} - \sum_{t=1}^{T} r_{a_t,t}\right], \tag{192}$$

where $g$ denotes a mapping from context set to available actions. For contextual bandits with non-adversarial rewards, Thompson sampling is a classical algorithm that achieves the state-of-the art performance in practice [24]. We express the ture reward generating distibution of context $s$ and action $a_t$ as $\nu_{s,a_t}$. We place a prior $\mu_{s,i,0}$ for a reward of context $s$ and action $i$. Then, this prior is updated to a posterior distibution. At each time step, Thompson sampling selects the action by

$$r_t \in \arg\max_{i=\{1,\ldots,K\}} \hat{r}_{i,t}, \quad \hat{r}_{i,t} \sim \mu_{s,i,t}. \tag{193}$$

Then corresponding posterior is updated by the observed reward.

Following the previous work [24, 21], we consider a neural network where the input is the context and the output is the $K$-dimensional, and we consider a prior on parameters of the network. We approximate the posterior of the neural network and express the uncertainty by the approximate posterior distribution. All the hyperparameters are exactly the same with the previous work [24].

In addition to the main paper, here, we also show the results of Var-svgd in Table 9.

# H Summary of the loss function based second-order Jensen inequalities

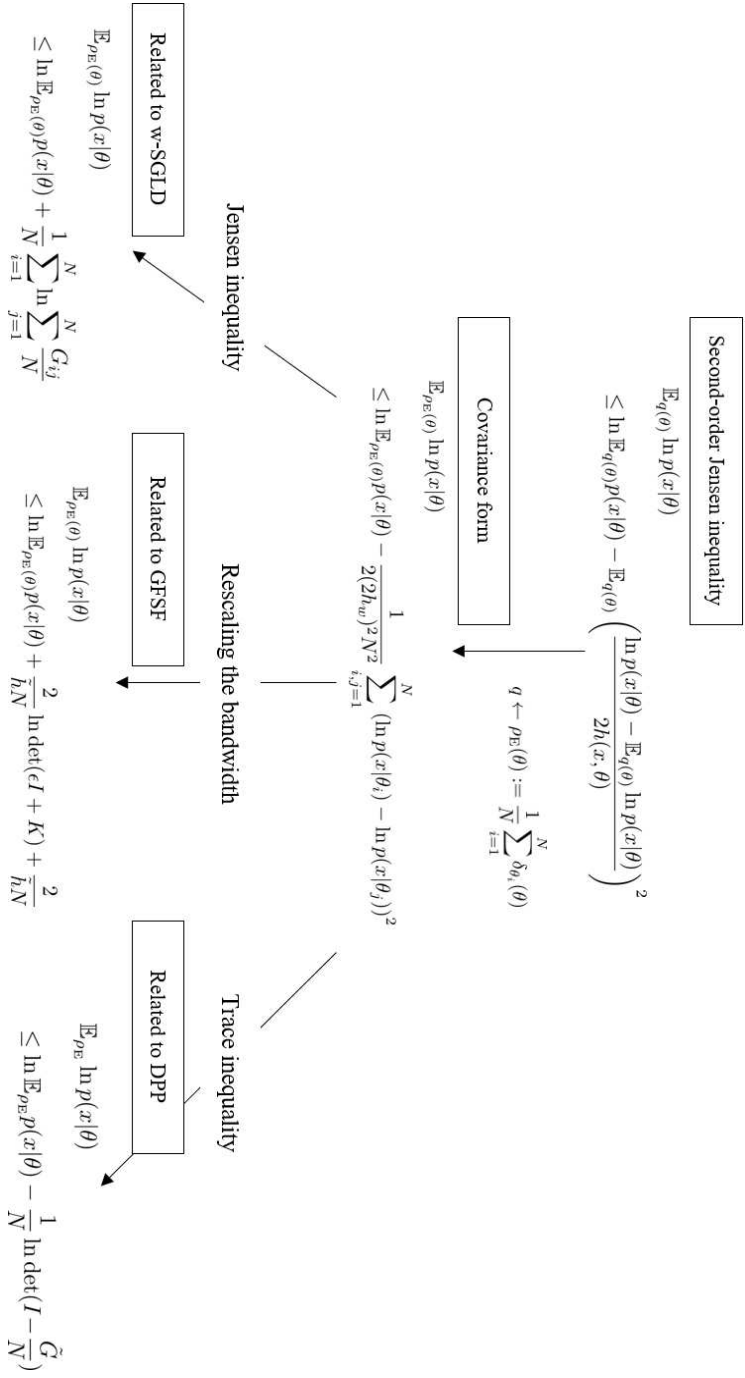

Figure 5: Summary of the second-order Jensen inequalities presented in this work.